\documentclass[letterpaper]{article} % DO NOT CHANGE THIS
\usepackage{aaai24}  % DO NOT CHANGE THIS
\usepackage{times}  % DO NOT CHANGE THIS
\usepackage{helvet}  % DO NOT CHANGE THIS
\usepackage{courier}  % DO NOT CHANGE THIS
\usepackage[hyphens]{url}  % DO NOT CHANGE THIS
\usepackage{graphicx} % DO NOT CHANGE THIS
\usepackage{natbib}  % DO NOT CHANGE THIS AND DO NOT ADD ANY OPTIONS TO IT
\usepackage{caption} % DO NOT CHANGE THIS AND DO NOT ADD ANY OPTIONS TO IT
\usepackage{algorithm}
\usepackage{algorithmic}

\usepackage{newfloat}
\usepackage{listings}
\DeclareCaptionStyle{ruled}{labelfont=normalfont,labelsep=colon,strut=off} % DO NOT CHANGE THIS
\floatstyle{ruled}
\newfloat{listing}{tb}{lst}{}
\floatname{listing}{Listing}
\usepackage{adjustbox}
\usepackage{booktabs}       % professional-quality tables
\usepackage{amsfonts}       % blackboard math symbols
\usepackage{nicefrac}       % compact symbols for 1/2, etc.
\usepackage{microtype}      % microtypography
\usepackage{xcolor}         % colors
\usepackage{amsmath}
\usepackage{amssymb}
\usepackage{amsmath}
\usepackage{mathtools}
\usepackage{multirow}
\usepackage{float}
\usepackage{bm}
\usepackage{graphicx}
\usepackage[para]{footmisc}
\usepackage{natbib}

\usepackage{caption}
\usepackage{subcaption}

\allowdisplaybreaks[4]

\definecolor{amber(sae/ece)}{rgb}{1.0, 0.49, 0.0}
\definecolor{cadmiumorange}{rgb}{0.93, 0.53, 0.18}
\definecolor{azure(colorwheel)}{rgb}{0.0, 0.5, 1.0}

\newtheorem{definition}{Definition}

\DeclareMathOperator*{\argmax}{arg\,max}
\DeclareMathOperator*{\argmin}{arg\,min}

\newcommand\hl[1]{\textbf{#1.}}
\newcommand\la{\leftarrow}

\def\ddefloop#1{\ifx\ddefloop#1\else\ddef{#1}\expandafter\ddefloop\fi}

\def\ddef#1{\expandafter\def\csname v#1\endcsname{\ensuremath{\boldsymbol{#1}}}}
\ddefloop ABCDEFGHIJKLMNOPQRSTUVWXYZabcdefghijklmnopqrstuvwxyz\ddefloop

\def\ddef#1{\expandafter\def\csname v#1\endcsname{\ensuremath{\boldsymbol{\csname #1\endcsname}}}}
\ddefloop {alpha}{beta}{gamma}{delta}{epsilon}{varepsilon}{zeta}{eta}{theta}{vartheta}{iota}{kappa}{lambda}{mu}{nu}{xi}{pi}{varpi}{rho}{varrho}{sigma}{varsigma}{tau}{upsilon}{phi}{varphi}{chi}{psi}{omega}{Gamma}{Delta}{Theta}{Lambda}{Xi}{Pi}{Sigma}{varSigma}{Upsilon}{Phi}{Psi}{Omega}{ell}\ddefloop

\def\ddef#1{\expandafter\def\csname bb#1\endcsname{\ensuremath{\mathbb{#1}}}}
\ddefloop ABCDEFGHIJKLMNOPQRSTUVWXYZ\ddefloop
\title{Controllable Expensive Multi-objective Learning \\ with Warm-starting Bayesian Optimization}

\author {
    Quang-Huy Nguyen\textsuperscript{\rm 1, \rm 2 \rm*},
    Long P. Hoang \textsuperscript{\rm 1 \rm*},
    Hoang V. Vu \textsuperscript{\rm 2},
    Dung D. Le \textsuperscript{\rm 1}
}
\affiliations {
    \textsuperscript{\rm 1} College of Engineering and Computer Science, VinUniversity, Vietnam\\
    \textsuperscript{\rm 2} FPT Software AI Center, Vietnam\\
    \{huy.nq, long.hp, dung.ld\}@vinuni.edu.vn, hoangvv18@fpt.com\\
    \textsuperscript{\rm *} These authors contributed equally to this research work
}

\begin{document}

\maketitle

\begin{abstract}
Pareto Set Learning (PSL) is a promising approach for approximating the entire Pareto front in multi-objective optimization (MOO) problems. However, existing derivative-free PSL methods are often unstable and inefficient, especially for expensive black-box MOO problems where objective function evaluations are costly to evaluate. In this work, we propose to address the instability and inefficiency of existing PSL methods with a novel controllable PSL method called Co-PSL. Particularly, Co-PSL consists of two stages: (1) \emph{warm-starting Bayesian optimization} to obtain quality priors Gaussian process; and (2) \emph{controllable Pareto set learning} to accurately acquire a parametric mapping from preferences to the corresponding Pareto solutions. The former is to help stabilize the PSL process and reduce the number of expensive function evaluations. The latter is to support real-time trade-off control between conflicting objectives. Performances across synthesis and real-world MOO problems showcase the effectiveness of our Co-PSL for expensive multi-objective optimization tasks. 
\end{abstract}

\section{Introduction}
\label{sec: intro}

Multi-objective optimization (MOO) demonstrates a wide range of fundamental and practical applications in various domains, from text-to-image generation \cite{lee2024parrot} to ejector design for fuel cell systems \cite{hou2024optimization}. However, real-world MOO problems often involve multiple conflicting and expensive-to-evaluate objectives. For example, recommendation systems need to balance accuracy and efficiency \cite{le2017indexable}, battery usage optimization requires trade-offs between performance and lifetime \cite{Attia2020ClosedloopOO}, robotic radiosurgery involves coordination of internal and external motion \cite{Schweikard2000RoboticMC}, and hyperparameter tuning faces the dilemma between generalization and computational time \cite{swersky2013multi}. Traditional MOO methods aim to find a finite set of Pareto optimal solutions that represent different trade-offs between the objectives and offer a snapshot of the feasible solutions. However, this approach becomes impractical as the number of objectives increases, since the number of solutions needed to cover the whole Pareto front grows exponentially, resulting in a high computational cost.

Pareto Set Learning (PSL) \cite{navon2021learning, hoang2023improving} is a promising approach that allows users to explore the entire Pareto front for MOO problems by learning a parametric mapping to align trade-off preference weights assigned to objectives with their corresponding Pareto optimal solutions. This approach allows users to make real-time adjustments among different objectives using the optimized mapping model.

\hl{Challenges} In this paper, we consider the problem of multi-objective optimization (MOO) over expensive-to-evaluate functions. In this setting, \cite{lin2022pareto} is the first published work that adopts the Pareto set learning approach using Bayesian optimization (BO) to effectively solve black-box expensive optimization problems by minimizing the number of function evaluations. This derivative-free optimization, however, is shown to be unstable and inefficient, leading to more expensive function evaluations.       

\hl{Approach} To address this issue, we propose a new PSL framework called Controllable Pareto Set Learning (Co-PSL) that consists of two phases:
\begin{itemize}
    \item \emph{Warm-starting Bayesian optimization:} We acquire the Gaussian Processes priors for the black-box functions through a method that approximates the Pareto front with an emphasis on diversity, known as DGEMO \cite{konakovic2020diversity}. %These priors play a crucial role in enhancing the stability of the subsequent PSL process and in diminishing the need for resource-intensive function evaluations.
    \item \emph{Controllable Pareto set learning:} encompasses learning a parameterized Hypernetwork \cite{ha2016hypernetworks} to construct a parametric mapping connecting preferences with the corresponding Pareto solutions. This enables trade-off control between conflicting objectives and supports real-time decision-making.
\end{itemize}

The obtained Gaussian processes prior to the warm-starting phase serve as an enhanced initial basis for the Pareto set model's learning process. In other words, this incorporation provides a more refined starting point, enabling the PSL model to better grasp the intricacies of the objective functions and facilitating a more effective exploration of the Pareto front.

\hl{Contribution and Organization.} Our contributions can be summarized in the following:
\begin{itemize}
    \item \textit{First}, we provide the mathematical formulation of the controllable Pareto set learning for expensive multi-objective optimization task (Section \ref{sec: PSL_EMOO})
    \item \textit{Second}, we introduce our Co-PSL with 2-stages: (i) warm-starting stage for learning prior Gaussian Process to adequately approximate the Pareto front with confidence quality (Section \ref{sec: warmstart}) and (ii) controllable parameter initialization to stabilize the Pareto Set model consistently and effectively (Section \ref{sec: parameter_init})
    \item \textit{Third}, we conduct comprehensive experiments on 6 synthesis and real-world multi-objective optimization problems to validate the effectiveness of our opposed Co-PSL compared to baseline methods (Section \ref{sec: experiments}).%on Expensive Multi-objective Optimization and Pareto Set Learning tasks. (Section \ref{sec: experiments})
\end{itemize}

\section{Preliminary}
\label{sec: preliminary}
\subsection{Expensive Multi-Objective Optimization}
We consider the following expensive continuous multi-objective optimization problem:

\begin{align}
    \label{eq: mop}
    \vx^* &= \argmin_{\vx \in \mathcal{X}} \ \vf(\vx), \\
    \vf(\vx) &= \big(f_1(\vx),f_2(\vx),\cdots, f_m(\vx)\big) \notag 
\end{align}    

where $\mathcal{X} \subset \mathbb{R}^n$ is the decision space, $f_i: \mathcal{X} \rightarrow \mathbb{R}$ is a black-box objective function. For a nontrivial problem, no single solution can optimize all objectives at the same time, and we have to make a trade-off among them. We have the following definitions for multi-objective optimization: 

\begin{definition}[Dominance]
    A solution $\vx^a$ is said to dominate another solution $\vx^b$ if $f_i(\vx^{a}) \leq f_i(\vx^b), \ \forall i \in \{1,...,m\}$ and $\vf(\vx^a) \neq \vf(\vx^b)$. We denote this relationship as $\vx^a \prec \vx^b$. % In addition,  $\vx^a$ is said to strictly dominate $\vx^b$, denoted by $\vx^a \prec_\text{strict} \vx^b$, if $f_i(\vx^{a}) < f_i(\vx^b), \  \forall i \in \{1,...,m\}$.
\end{definition}

\begin{definition}[Pareto Optimality]
A solution $\vx^*$ is called Pareto optimal solution if $\nexists \vx^b \in \mathcal{X}: \ \vx^b \prec \vx^*$.  % Similarly, $\vx^\circ$ is called weakly Pareto optimal solution if $\nexists \vx^b, \ \vx^b \prec_\text{strict} \vx^\circ$.
\end{definition}

\begin{definition}[Pareto Set/Front]
The set of Pareto optimal is Pareto set, denoted by $\mathcal{P} = \{\vx^*\} \subseteq \mathcal X$, and the corresponding images in objectives space are Pareto front $\mathcal{P}_f = \{\vf(\vx) \mid \vx \in \mathcal{P}\}$.
\end{definition}

\begin{definition}[Hypervolume]
    Hypervolume \cite{zitzler1999multiobjective} is the area dominated by the Pareto front. Therefore the quality of a Pareto front is proportional to its hypervolume. Given a set of $n$ points $\vy = \{y^{(i)} | y^{(i)} \in \mathbb{R}^m; i=1,\dots, n\}$ and a reference point $\rho \in\mathbb{R}^m$, the Hypervolume of $\vy$ is measured by the region of non-dominated points bounded above by $y^{(i)} \in \vy$, then the hypervolume metric is defined as follows:
    \begin{equation}
        HV(\vy) = VOL\left(\underset{y^{(i)} \in \vy, y^{(i)} \prec \rho}{\bigcup}\displaystyle{\Pi_{i=1}^n}\left[y^{(i)},\rho_i\right]\right)
    \end{equation}

    where $\rho_i$ is the i$^{th}$ coordinate of the reference point $\rho$ and $\Pi_{i=1}^n\left[y^{(i)},\rho_i\right]$ is the operator creating the n-dimensional hypercube from the ranges $\left[y^{(i)},\rho_i\right]$.
    
\end{definition}

\subsection{Gaussian Process and Bayesian Optimization }
A Gaussian Process with a single objective is characterized by a prior distribution defined over the function space as:
\begin{eqnarray}
f(\vx) \sim GP(\mu(\vx),k(\vx,\vx)),
\label{eq: gaussian_process}
\end{eqnarray}
where $\mu: \mathcal{X} \rightarrow \bbR$ represents the mean function, and $k: \mathcal{X} \times \mathcal{X} \rightarrow \bbR^2$ is the covariance kernel function. Given $n$ evaluated solutions $\vD = \{\vX, \vy\} = \{(\vx^{(i)},f(\vx^{(i)})|i = 1,\ldots,n)\}$,  the posterior distribution can be updated by maximizing the marginal likelihood based on the available data. For a new solution $\vx^{n+1}$, the posterior mean and variance are given by:
\begin{align*}
% \label{eq: gaussian_process_prediction}
\hat{\mu}(\vx^{(n+1)}) &= \vk^T\vK^{-1}\vy, \\
\hat{\sigma}^2(\vx^{(n+1)}) &= k(\vx^{(n+1)},\vx^{(n+1)}) -  \vk^T \vK^{-1}\vk, 
\end{align*}
where $\vk = k(\vx,\vX)$ is the kernel vector and $\vK = k(\vX, \vX)$ is the kernel matrix.

Bayesian optimization involves searching for the global optimum of a black-box function $f(\cdot)$ by wisely choosing the next evaluation via the current Gaussian process and acquisition functions. A new evaluation $\vx^{n+1}$ is determined through an acquisition function $\alpha$, which guides the search for the optimal solution. More specifically, the next evaluation $\vx^{n+1}$ is selected as the optimal solution of the acquisition in order to make the best improvement:
\begin{eqnarray}
\vx^{n+1} = \underset{x}{\argmax}\ \alpha(\vx; \vD),
\label{eq: acquisition_funcition}
\end{eqnarray}
Common choices for the acquisition function include the upper confidence bound (UCB), expected improvement (EI), and Thomson sampling (TS). We would like to refer the readers to \cite{frazier2018tutorial} for a detailed introduction. 

Additionally, in the setting of multi-objective optimization, hypervolume is a common criterion for determining the quality of the Pareto front. For multi-objective Bayesian optimization, we also use Hypervolume Improvement (HVI) as the additional acquisition function for selecting new solutions.

\begin{definition}[Hypervolume Improvement]
    Hypervolume Improvement (HVI) determines how much the Hypervolume would increase if a set of new $\vX_+$ is added to the current dataset $\{\vX, \ \vY\}$:
    \begin{equation*}
        HVI\big(\vf(\vX_+),\ \vY\big) = HV\big(\vY \cup \vf(\vX_+)\big) - HV(\vY)
    \end{equation*}
\end{definition}

\subsection{Pareto Set Learning}
\label{sec: PSL_EMOO}
Pareto Set Learning approximates the entire Pareto Front of Problem (\ref{eq: mop}) by directly approximating the mapping between an arbitrary preference vector $r$ and a corresponding Pareto optimal solution computed by surrogate models: % (which can be Expected Improvement (EI), Upper Confidence Bound (UCB), or :
\begin{align}
    \label{eq: PSL}
    & \theta^* = \argmin_{\theta} \mathbb{E}_{r \sim \text{Dir}(\alpha)} g(\hat{\vf}(\vx_r)| r) \\
    & \text{s.t } \ \vx_r = h(r|\ \theta) \in \mathcal{P}, \ h(\bbS^m|\ \theta^*) = \mathcal{P} \nonumber
\end{align}
where $\text{Dir}(\alpha)$ is the flat Dirichlet distribution with $\alpha = \left(\frac{1}{m}, \dots, \frac{1}{m} \right) \in \mathbb{R}^m$, $\bbS^m = \{ r \in \mathbb{R}^m_{>0}: \sum_i r_i = 1\}$ is the feasible space of preference vectors $r$, $\hat{f}_i(\cdot): \mathcal{X} \rightarrow \mathbb{R}, \forall i \in \{1,\dots,m\}$ are surrogate models, $\hat{\vf}(\cdot) = \left[ \hat{f}_i(\cdot) \right]_{i=1}^m$, the scalarization function  $g: \mathbb{R}^m \times \bbS^m \rightarrow \mathbb{R}$ helps us map a given preference vector with a Pareto solution, and $h (\cdot, \cdot): \bbS^m \times \Theta \rightarrow \mathcal{X}$ is called the Pareto set model, approximating the mentioned mapping.

There are many options available in the field of surrogate models, such as Expected Improvement (EI), Upper Confidence Bound (UCB), and Lower Confidence Bound (LCB). In the scope of this study, we choose to use LCB as our surrogate model, which is:
\begin{align}
    \vf_i(\vx) = \hat{\mu}(\vx) -  \lambda \hat{\sigma}(\vx)
\end{align}
The landscape provides various choices for scalarization functions like Linear Scalarization, Chebyshev, and Inverse Utility. However, we decided to use the Chebyshev function, 

\begin{align}
    g(\hat{\vf}(\vx)| r) = \max_i \{r_i \lvert\hat{f}_i(\vx) - z_i^*\rvert \} \ \  \forall i \in \{1, \dots, m \}
\end{align}

where $z^*$ is an ideal objective vector. This choice for scalarization is supported by the Pareto Front Learning context's exceptional stability, particularly when dealing with non-convex known functions as shown in \cite{tuan2023framework}. Significantly, the intersection of a preference vector and the Pareto Front $\mathcal{P}_f$ represents the image of the corresponding Pareto optimal solution in the objective space.

In truth, obtaining a complete and accurate approximation of the entire variable space using surrogate models is impossible. However, the essence of Pareto set learning lies in the precise approximation of the feasible optimal space. This space, existing as a continuous manifold, represents a distinct subset within the broader variable space. Through a well-considered strategy, Pareto Set Learning demonstrates its ability to establish a highly accurate mapping between the preference vector space and the Pareto continuous manifold. However, optimizing the Pareto Set Model can be challenging and unstable if Gaussian processes do not approximate well black-box functions.

\section{Related work}
\label{sec: related}
{\bf Multi-objective Bayesian Optimization.} 
% \label{sec: related_MOBO}
Conventional Multi-objective Bayesian optimization (MOBO) methods have primarily concentrated on locating singular or limited sets of solutions. To achieve a diverse array of solutions catering to varied preferences, scalarization functions have emerged as a prevalent approach. Notably, \cite{paria2020flexible} adopts a strategy of scalarizing the Multi-objective problem into a series of single-objective ones, integrating random preference vectors during optimization to yield a collection of diverse solutions. Meanwhile, \cite{abdolshah2019multiobjective} delves into distinct regions on the Pareto Front through preference-order constraints, grounded in the Pareto Stationary equation.

Alternatively, evolutionary and genetic algorithms have also played a role in furnishing diversified solution sets, as observed in \cite{zhang2009expensive} which concurrently tackles a range of surrogate scalarized subproblems within the MOEA/D framework \cite{zhang2007moea}. Additionally,  \cite{bradford2018efficient} adeptly combines Thompson Sampling with Hypervolume Improvement to facilitate the selection of successive candidates. Complementing these approaches, \cite{konakovic2020diversity} employs a well-crafted local search strategy coupled with a specialized mechanism to actively encourage the exploration of diverse solutions. A notable departure from conventional methods, \cite{belakaria2020uncertainty} introduces an innovative uncertainty-aware search framework, orchestrating the optimization of input selection through surrogate models. This framework efficiently addresses MOBO problems by discerning and scrutinizing promising candidates based on measures of uncertainty.

\begin{figure*}[!ht]
    \centering
    \begin{subfigure}[b]{0.48\textwidth}
        \centering
        \includegraphics[width=\textwidth]{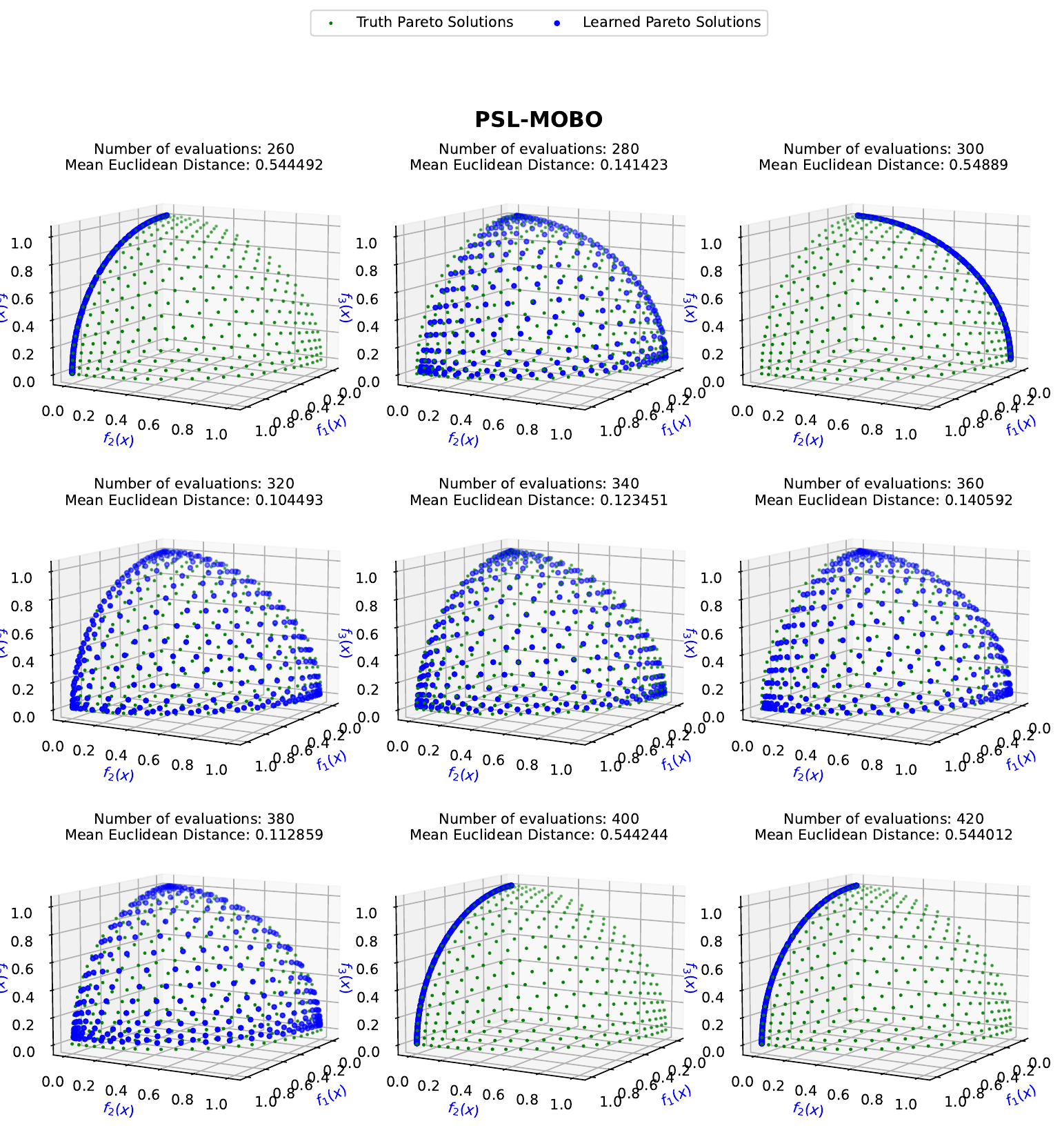}
    \end{subfigure}
    \begin{subfigure}[b]{0.48\textwidth}
        \centering
        \includegraphics[width=\textwidth]{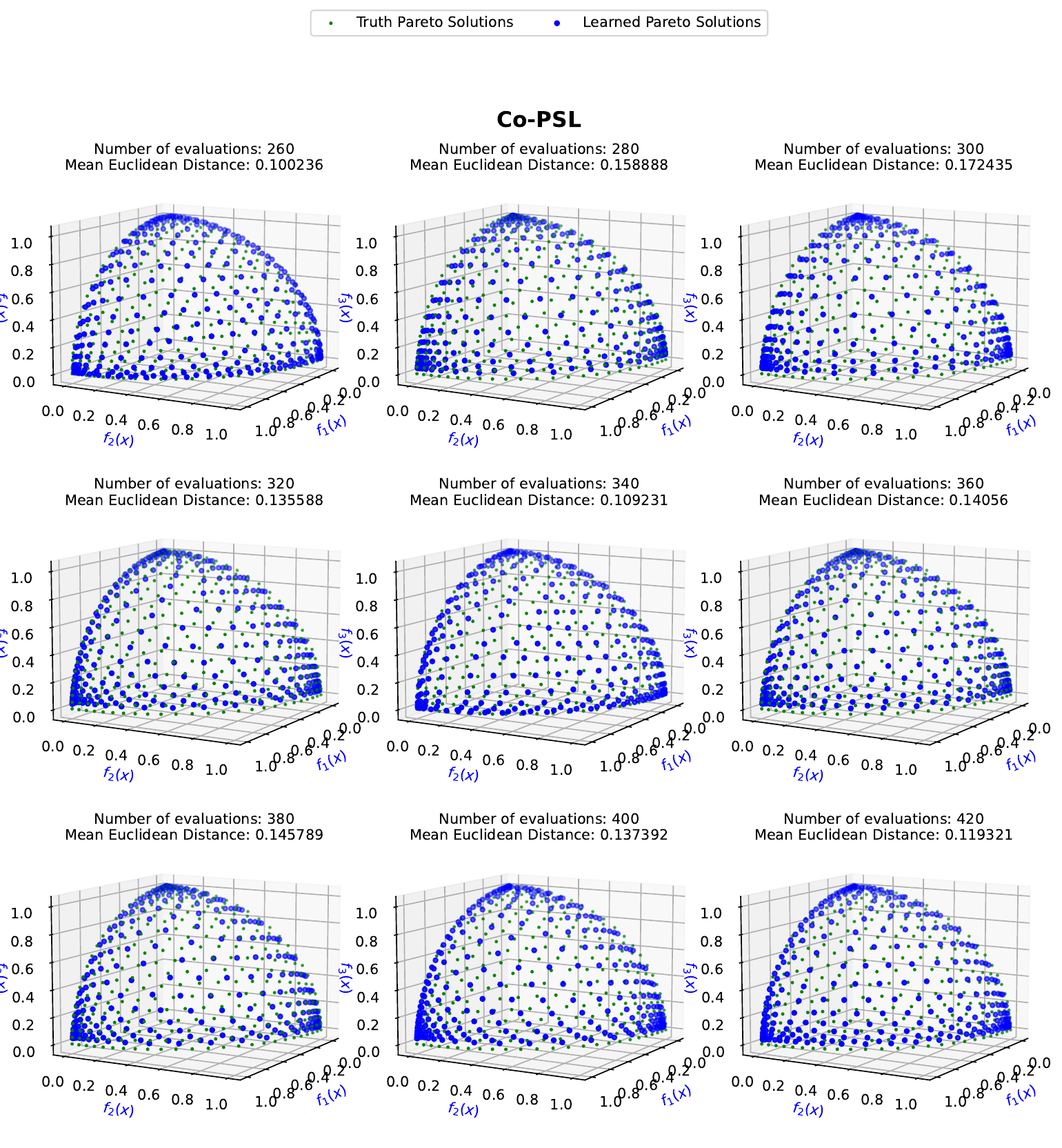}
    \end{subfigure}
    \caption{Performance of traditional Pareto Set Learning (PSL-MOBO) and our Co-PSL on DTLZ2 problem. We use the Mean Euclidean Distance between truth and estimated Pareto solutions under the same set of reference vectors to determine the performance of the Pareto Set Model, and thus, the quality of the learned Pareto Front. While the main baseline PSL-MOBO exhibits a highly unstable and uncertain performance in obtaining the Pareto front, Co-PSL demonstrates a smoother and more consistent front across optimization iterations.}
    \label{fig: unstable_PSL}
\end{figure*}

{\bf Pareto Set Learning}
\label{sec: related_PSL}
or PSL emerges as a proficient means of approximating the complete Pareto front, the set of optimal solutions in multi-objective optimization (MOO) problems, employing Hypernetworks \cite{chauhan2023brief}. This involves the utilization of hypernetworks to estimate the relationship between arbitrary preference vectors and their corresponding Pareto optimal solutions. In the realm of known functions, effective solutions for MOO problems have been demonstrated by \cite{navon2021learning, hoang2023improving, tuan2023framework} while in the context of combinatorial optimization, \cite{lin2022pareto_com} have made notable contributions. Notably, \cite{lin2022pareto} introduced a pioneering approach to employing PSL in the optimization of multiple black-box functions, PSL-MOBO, leveraging surrogate models to learn preference mapping, making it the pioneer method in exploring Pareto Set learning for the black-box multi-objective optimization task. However, PSL-MOBO is hindered by its reliance on Gaussian processes for training the Pareto set model, leading to issues of inadequacy and instability in learning the Pareto front.

\paragraph{Challenge of Pareto Set Learning for Multi-objective Bayesian Optimization.}
In this section, we analyze the instability of the Gaussian process when conducting Bayesian optimization for Pareto Set learning and propose our method, Co-PSL, to mitigate the challenge. Conducting Bayesian optimization on a black-box function $f(x)$, via the surrogate Gaussian Process will result in efficient solutions $\{x, f(x)\}$ towards the optimal solution $f(x^*)$. However, in the process of seeking the optimal solution, the surrogate model will be broken down into many small uncertain regions between the explored solutions, and under LCB as the acquisition function, these regions will become ``pseudo-local optimals'' that make the quest for the optimal solution challenging.

On the other hand, Pareto Set Learning seeks to learn the mapping between the trade-off reference vector $r$ and its corresponding truth Pareto solution $\hat{\vf}(x_r)$ through the Pareto Set Model $x = h(r| \theta)$. In PSL-MOBO, at the optimization iteration $i$-th, given a set of prior solutions $\{\vx_{i-1}, \vf(x)_{i-1}\}$, the solutions learned by the Pareto Set Model $h(r|\theta)$ are initialized around a fixed nadir value $0$ at the beginning of each iteration, then moved to the optimal region determined by LCB through a small training loop of $T$ steps. However, with the existence of these pseudo-local optimals, the computed solution can be stuck in any of these regions, making $h(r|\theta)$ fail to reach the optimal regions for effective mapping. This resulted in the instability of Pareto Set Learning in PSL-MOBO, as depicted in Figure \ref{fig: unstable_PSL}.

\section{Controllable Pareto Set Learning}
\label{sec: method}
To deal with the instability when solving the MOBO problem based on the Gaussian Process, we proposed Controllable Pareto Set Learning (Co-PSL) with two main improvements: (i) warm-starting Bayesian Optimization (Section \ref{sec: warmstart}) to obtain a good approximation of individual black-box functions and thus mitigate the existence of uncertainty regions that leads to unstable Pareto Set Learning, and (ii) improving Pareto Set Learning with parameter re-initialization (Section \ref{sec: parameter_init}) between optimizing iterations to support the Pareto Set Model passing through these uncertainty regions for effectively black-box optimization. A detailed description of how Co-PSL operates is presented on Section \ref{sec: CoPSL} and Algorithm \ref{algo: algorithm}.

\subsection{Warm-starting Bayesian Optimization}
\label{sec: warmstart}

In this section, we describe our approach to leveraging warm-starting Bayesian optimization to improve both the instability when training the Pareto Set Model and the performance of the learned Pareto Front. We argue that having an un-approximate Gaussian process with high uncertainty regions, or ``pseudo-local optimal'', might not favor black-box MOO in general and Pareto Set Learning in particular. Thus, obtaining a good approximation at first is a necessary step. Therefore, we perform a warm-starting Bayesian optimization solely on the individual black-box function $f(x)$ to approximate $f(x)$ as a prior surrogate model, obtaining a good prior Pareto front for the following stage. This comes with two benefits: (1) the approximated surrogate model will lead to a reliable Pareto front, helping the Pareto Set Model learn the mapping effectively and confidently; and (2) the well-approximate surrogate model will multigate the size and range of these ``pseudo-optimal'', supporting the latter Pareto Set Learning phase with smoother and more consistent optimizing performance. 

For the warm-starting Bayesian optimization, we adapted DGEMO \cite{konakovic2020diversity} for approximate $\vf(x)$ with a batch size of $b_1$. The algorithm, which samples a set of diverse evaluations across different regions of the estimated Pareto front with the highest HVI, is an ideal choice since it produces a set of diverse solutions that both approximate individual $f(x)$ overall and obtain a reliable prior Pareto front comprehensively.

\begin{algorithm}[t!]
\caption{Co-PSL main algorithm}
\label{algo: algorithm}
\textbf{Input}: Black-box multi-objective objectives $\vf(\vx) = \{f_j(\vx),\ j \in 1, \cdots, m\}$ and initial evaluation samples $\big\{\vx_0, \ \vf(\vx_0)\big\}$\\
\textbf{Main Parameter}: $n_1, \ n_2$: number of iteration for stage 1 and stage 2 respectively; $b_1, \ b_2$: batch evaluations for stage 1 and stage 2 respectively; $S_m$: feasible preference vector space on $\bbR^m$; $\theta$: parameter for Pareto set model $h$; $\beta$: positive scalar.\\
\begin{algorithmic}[1] %[1] enables line numbers
\STATE $\vD \la \big\{\vx_0, \ \vf(\vx_0)\big\} $
\STATE \textit{\# Stage 1: Warm-starting the Gaussian Process prior}
\FOR{i $\la 0$ to $n_1$}
\STATE Training GP $\hat{f}_j(\vx)$ for each $f_j(\vx)$ on $\vD$,\\ $\hat{\vf}_j(\vx) \la \big\{\hat{f}_j(\vx)\big\}$
\STATE Approximating the Pareto front $\mathcal{P}$ over $\hat{\vf}(\vx)$ 
\STATE Selecting subset $\{\vx\}_{b_1} \in \mathcal{X}$ that has highest HVI 
\STATE $\vD \la \vD \cup \big\{\vx, \ \vf(\vx)\big\}_{b_1}$
\ENDFOR
\STATE \textit{\# Stage 2: Controllable Pareto Set Learning}
\STATE Initialize Pareto set model $h(r|\ \theta_0)$
\FOR{i $\la 0$ to $n_2$}
\STATE Training GPs on $\vD$ and approximating $\mathcal{P}$
\IF{$i > 0$}
\STATE Intialize $\theta_i$ by Formula (\ref{eq: theta_update})
\ENDIF
\FOR{t $\la 0$ to $T$}
\STATE Randomly sample $K$ vectors $\{r^k\}_{k=1}^K \sim \bbS^m$
\STATE Update $\theta_i$ with gradient descent by Formula (\ref{eq: gp_theta}) to optimize Formula (\ref{eq: PSL})
\ENDFOR
\STATE Randomly sample $B$ vectors $\{r^b\}_{b=1}^B \sim \bbS^m$, then compute $\vX_{r^b} = h(r^b|\ \theta)$.
\STATE Selecting subset $\{\vx\}_{b_2} \in \vx_{r^b}$ that has highest HVI
\STATE $\vD \la \vD \cup \big\{\vx, \ \vf(\vx)\big\}_{b_2}$
\ENDFOR
\end{algorithmic}
\textbf{Output}: Total evaluated solutions $\vD = \big\{\vx,\ \vf(\vx)\big\}$ and the final parameterized Pareto set model $h(r|\ \theta)$
\end{algorithm}

\subsection{Paremeter initialization for Pareto Set Learning}
\label{sec: parameter_init}

By initializing the solution around $0$ and gradually moving to the optimal region by learning from an updated set of solutions for each iteration, the Pareto Set Model $h(r|\theta)$ can easily fail and get stuck in these ``pseudo-local optimal'' resulted in GPs that leads to instable performance. To tackle this issue, we re-initialize its parameters $\theta$ at iteration $i$-th to effectively overcome these regions based on the performance of previous steps, as follows:

\begin{equation}
   \theta_i = \beta\cdot \theta_{i-1} + (1-\beta)\cdot G(\theta_i)
   \label{eq: theta_update}
\end{equation}

where $G$ is the random noise initialization and $\beta$ is a sufficiently small positive scalar as a trade-off parameter. For $G$, we use Glorot initialization based on Uniform distribution \cite{glorot2010understanding}. For $\beta$, we start with $\beta = 0.2$ and reduce by a factor of $10$ for every $5$ training iteration. With this parameter initialization scheme, the Pareto set model $h(r|\theta)$ can reduce the chance of being stuck at pseudo-optimal as it is based on the stopping position of the previous iteration. Optimizing $h(r|\theta)$ under scalarization function $g$ can be done with learning rate $\eta$ at the learning step $t$ as follows:

\begin{equation}
    \theta_{t} = \theta_{t-1} - \eta \nabla_\theta g\Big(\hat\vf\big(\vx_r = h(r|\theta_{t-1})\big)\Big|r\Big)
    \label{eq: gp_theta}
\end{equation}

\subsection{Optimizing Co-PSL}
\label{sec: CoPSL}
For each optimizing iteration, we trained the Pareto Set Model $h(r|\theta)$ with $T$ training step, whereas we randomly sampled $K$ preference $\{r^k\}_{k=1}^K \sim \text{Dir}(\alpha), \alpha \in \mathbb{R}^m$ for optimizing $\theta$. Then, we sampled $B$ preference $\{r^b\}_{b=1}^B \sim \text{Dir}(\alpha)$ to compute the corresponding evaluation $\vx_{r^b}$. Finally, we selected a subset $\{\vx\}_{b^2} \in \{\vx_{r^b}\}_{b=1}^B$ including $b_2$ elements that achieve the highest HVI for the next batch of expensive evaluations:
\begin{align}
    \{\vx\}_{b^2} & = \argmax_{\mathbf{X}_+} HVI(\hat\vf(\mathbf{x}_+, D_y) \\
    & \text{s.t } \left| \mathbf{x}_+ \right| = b_2, \ \mathbf{x}_+ \in \{\vx_{r^b}\}_{b=1}^B \notag
\end{align}

\section{Experiments}
\label{sec: experiments}

\subsection{Experiment setup}

\begin{figure*}[ht!]
    \centering
    \includegraphics[width = \textwidth]{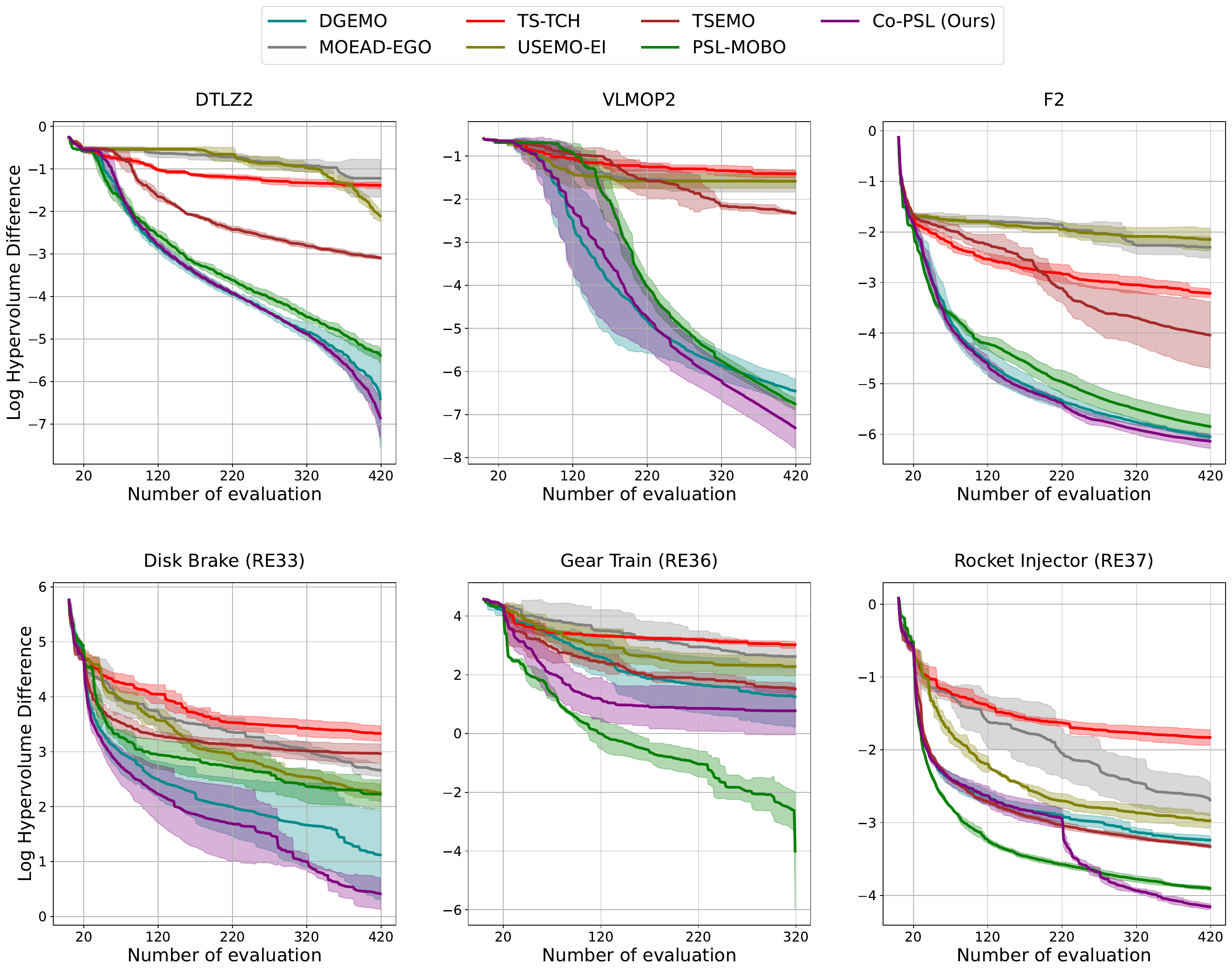}
    \caption{Mean Log Hypervolume Differences between the truth Pareto Front and the learned Pareto Front with respect to the number of expensive evaluations on all MOBO algorithms, with synthesis problems on the top row and real-world problems on the bottom row. The solid line is the mean value averaged among 5 independent runs, and the shaded region is the standard deviation around the mean value.}
    \label{fig: comparision_LSD}
\end{figure*}

\hl{Evaluation Metrics} The area dominated by the Pareto front is known as Hypervolume. The higher the Hypervolume, the better the Pareto front quality. We propose two metrics for benchmarking the quality of the learned Pareto front and the Pareto set model. For evaluating the quality of the learned Pareto front, we employ Log Hypervolume Difference (LHD) between the Hypervolumes computed by the truth Pareto front $\mathcal{P}_f$ and the learned Pareto front $\hat{\mathcal{P}}_f$ as follows:

\begin{equation*}
LHV(\mathcal{P}_f, \hat{\mathcal{P}}_f) = \log\Big(\text{HV}\big(\mathcal{P}_f\big) - \text{HV}\big(\hat{\mathcal{P}}_f\big)\Big)    
\end{equation*}

For evaluating the quality of the Pareto set model, mapping preferences to the corresponding Pareto solutions, given a set of evenly distributed preferences $\mathbf{r} = \{r^k\}_{k=1}^K$, we employ Mean Euclidean Distance (MED) \cite{tuan2023framework} between truth corresponding Pareto optimal solutions $\mathcal{P} = \{ \vf(\vx^*_{r^k}), r^k \in \mathbf{r}\}$ and learned solutions $\hat{\mathcal{P}} = \{ \hat{\mu}(h(r^k|\theta)), r^k \in \mathbf{r}\}$ as follows:
\begin{equation*}
MED(\mathcal{P}, \hat{\mathcal{P}}) = \frac{1}{K}\sum_{k=1}^K\Big(\big\lVert \vf(\vx^*_{r^k}) - \hat{\mu}(h(r^k|\theta)) \big\rVert_2\Big)   
\end{equation*}

with $K = 200$ for 2-objective problems and $K = 300$ for 3-objective problems.
 
The lower the MED and LHB scores, the better the performance of the MOBO method in evaluations. In addition, by drawing MED plots during the training process, we can evaluate the stability of the Pareto set model when dealing with black-box multi-objective optimization problems.

\hl{Multi-Objective Problems} For method evaluations, we employed three synthesis problems and three practical real-world problems. The three synthesis problems includes VLMOP2 \cite{van1999multiobjective}, DTLZ2 \cite{deb2002scalable}, and F2 \cite{lin2022pareto}. The three real-world problems include Disk Brake Design (RE33) \cite{ray2002swarm}, Gear Train Design (RE36) \cite{deb2006innovization}, Rocket Injector Design (RE37) \cite{vaidyanathan2003cfd}. Detailed descriptions of these problems are presented in the Appendix. As the three real-world problems do not have the exact Pareto Front, we used the approximation front proposed by \cite{tanabe2020easy} as the replacements. While all problems can be evaluated under the LHD, only 3 synthesis problems can be evaluated under MED.

\hl{Baselines} We compare our proposed Co-PSL with current state-of-the-art of MOBO PFL methods including TS-TCH \cite{paria2020flexible}, USeMO-EI \cite{belakaria2020uncertainty}, MOEA/D-EGO \cite{zhang2009expensive}, TSEMO \cite{bradford2018efficient}, DGEMO \cite{konakovic2020diversity}, and PSL-MOBO \cite{lin2022pareto}. All of the methods can be evaluated under the LHD score with respect to the number of expensive evaluations throughout the training. However, only PSL-MOBO can be evaluated under MED as the algorithm trains the Pareto set model for computing the Pareto solution based on input trade-off preference.

\hl{Training Setting} For the warm-starting stage, we start with 20 random initial evaluations, and train the prior Gaussian Process on $n_1 = 20$ iterations with a batch evaluation of $b_1 = 10$, making a total of 220 expensive evaluations. In the second controllable Pareto Set Learning stage, we trained with $n_2 = 20$ iterations with a batch evaluation of $b_2 = 10$, resulting in 200 expensive evaluations. Thus, Co-PSL is trained with a total of 40 iterations and 420 expensive evaluations, with the number of evaluations divided equally among the two stages. For other baseline MOBO methods, we start with 20 random evaluations and train by 40 iterations with a batch evaluation of 10, resulting in the same number of evaluations. The only exception is Gear Train Design (RE36), where we set $n_2 = 10$ as the problem's variables accept integer values only, resulting in concave and disconnected Pareto front \cite{tanabe2020easy}. Reducing the PSL training stage prevents the Pareto set model from having redundant expensive evaluations, resulting in an overfitted Pareto front that prevents effective method evaluation.

\begin{figure}[hb!]
    \centering
    \includegraphics[width = \linewidth]{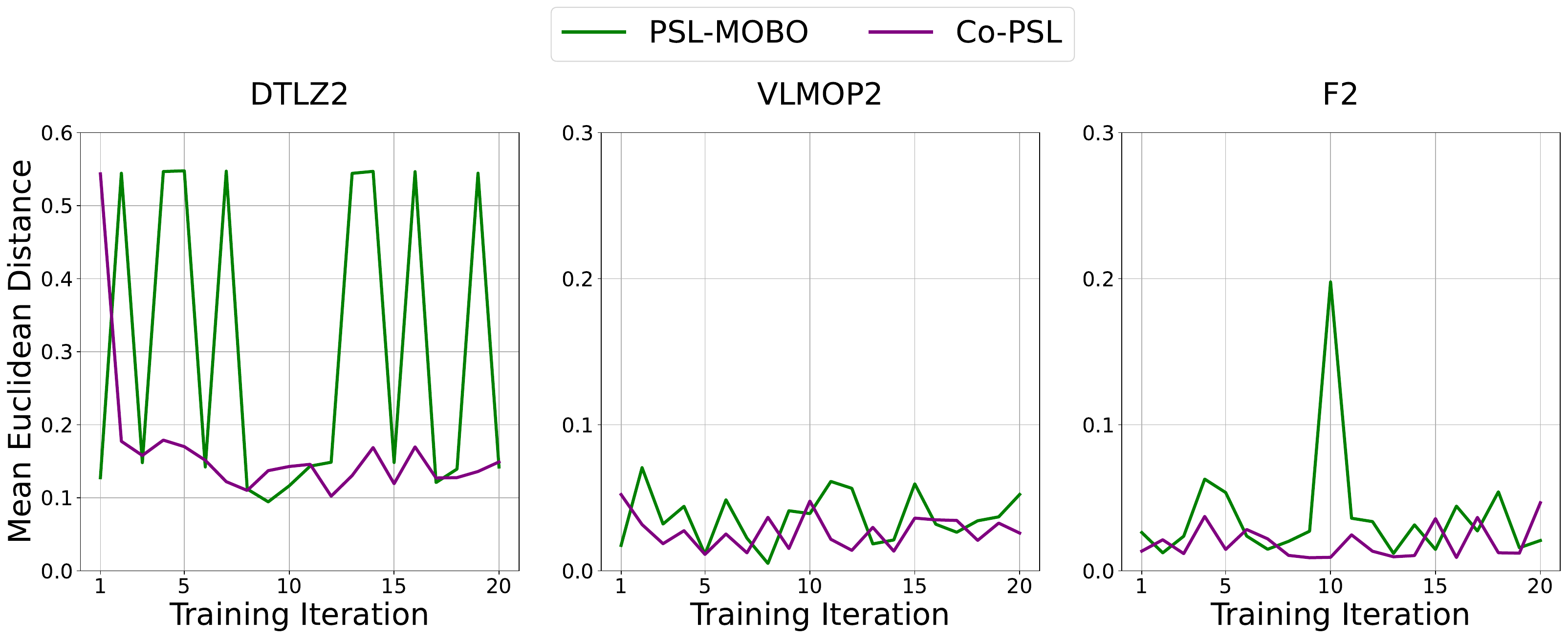}
    \caption{Mean Euclidean Distance between the truth Pareto solutions and the predicted Pareto solutions under the same set of preference vectors with respect to the number of training iterations between PSL-MOBO (baseline) and Co-PSL (ours) among the three synthesis problems.}
    \label{fig: comparision_MED}
\end{figure}

\begin{figure}[ht!]%
    \centering
    \subfloat[][\centering Trade-off performances between 2 stages of Co-PSL]{\includegraphics[width=0.45\textwidth]{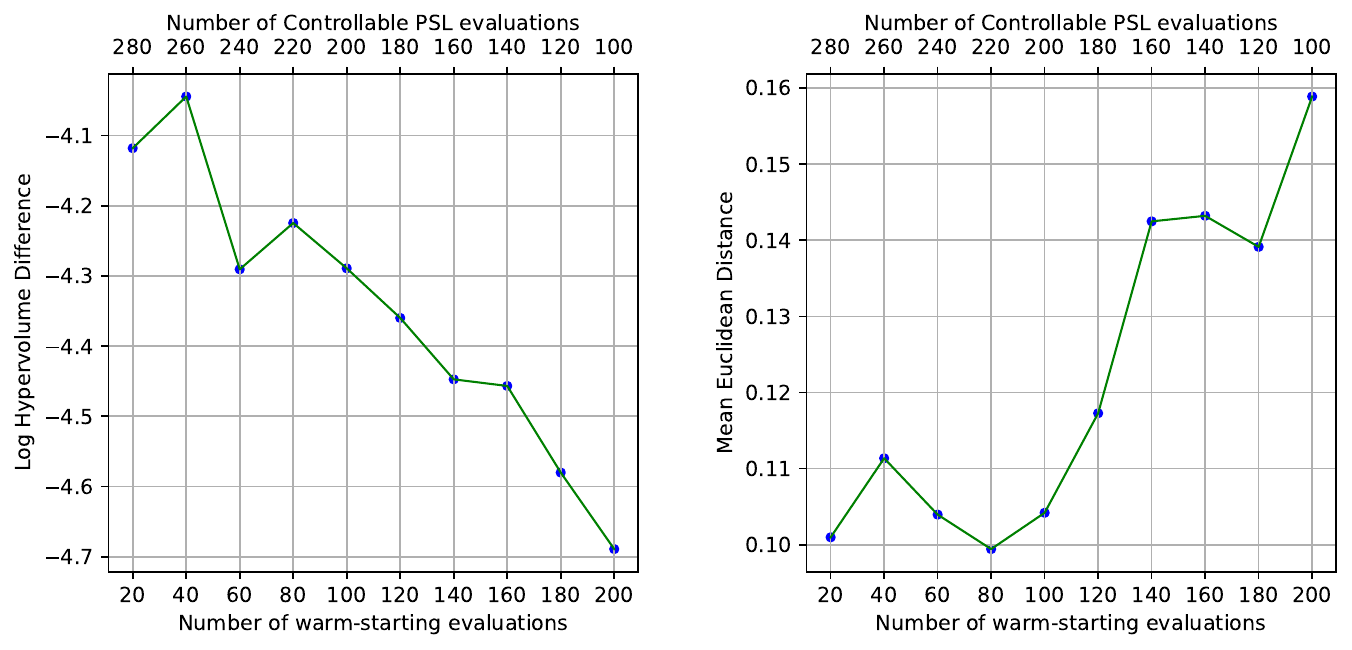} \label{fig: Co_PSL_tradeoff_2stage}}
    \quad
    \subfloat[][\centering Trade-off performances with different $\beta$]{\includegraphics[width=0.45\textwidth]{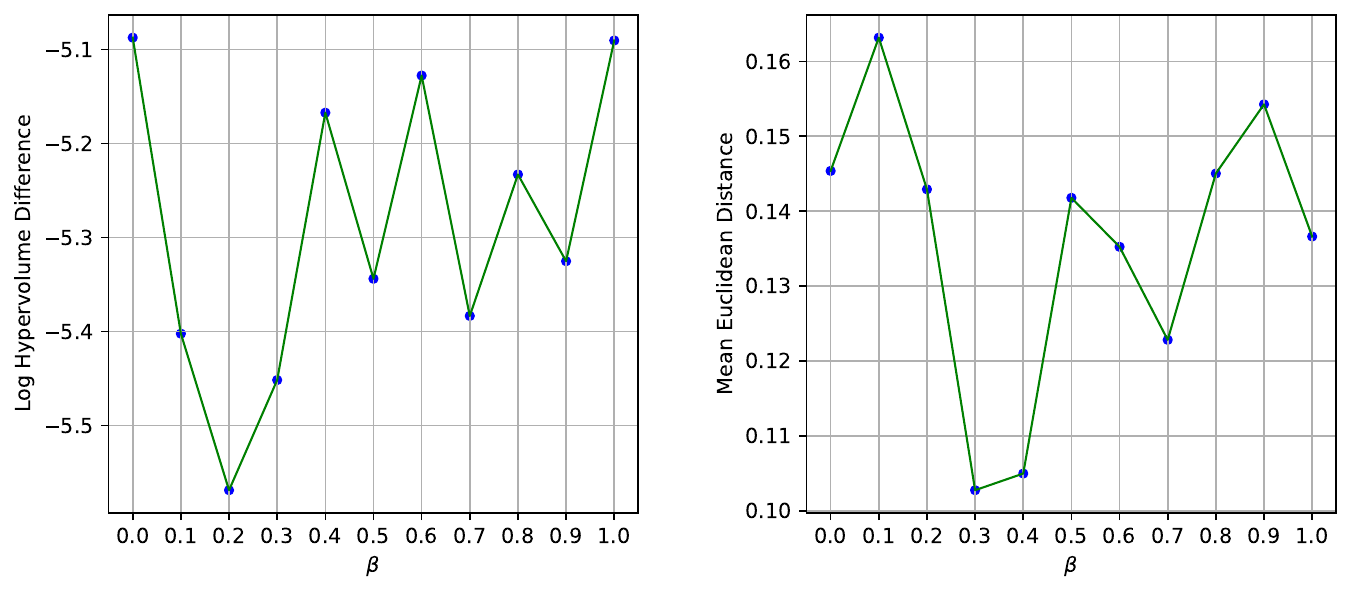}\label{fig: Co_PSL_tradeoff_beta}}
    \caption{Ablation studies for Co-PSL on DTLZ2 with Log Hypervolume Difference (left column) and Mean Euclidean Distance (right column), including (a) trade-off between warm-starting and Controllable Pareto Front Learning and (b) trade-off $\beta$ between previous parameter and random initialization for Pareto Set Model parameter initialization.}
\end{figure}

\subsection{Results and Analysis}

\hl{Performance of expensive MOO task}
Figure \ref{fig: comparision_LSD} presents the mean Log Hypervolume Difference between the truth Pareto Front and the learned Pareto Front across expensive evaluations on six testing problems, with the solid lines as the mean values averaged over five independent runs and the shaded regions as standard deviations. Overall, our Co-PSL outperforms all baseline methods on most problems, with the only exception of RE36 (Gear Train Design), where our method is surpassed by PSL-MOBO. However, in that case, we recognized that the Pareto front of the problem is concave and disconnected, with the variables being integer only, leaving the problem extremely challenging. As the first stage of Co-PSL derived directly from DGEMO, we recognized that the first half regions between the two methods nearly overlapped with each other, especially on DTLZ2, F2, and RE37. Furthermore, both two methods are recognized to have large standard deviations on RE33, RE36, and VLMOP2.

\hl{Performance of Pareto Set Learning task}
Figure \ref{fig: comparision_MED} illustrates the Mean Euclidean Distance between the computed Pareto solutions and the truth Pareto solution under the same set of evenly distributed preference vectors $\{r^k\}_{k=1}^K$ across PSL training iterations between PSL-MOBO \cite{lin2022pareto} and Co-PSL. For a fair comparison, we only employ the MED results from the last 20 training iterations of PSL-MOBO respecting the 20 training iterations of Co-PSL, sharing similar amounts of expensive evaluations to approximate the Pareto front. Our Co-PSL achieved better and more consistent MED score plots compared to the baseline method, especially in DTLZ2. In that 3-objective problem, PSL-MOBO is recognized to be highly inconsistent and unstable across iterations, whereas our Pareto Set Model is more stable and performs better under MED scores during training. In the other two synthesis 2-objective problems, both methods have more stable MED scores. Therefore, the baseline method becomes unstable when dealing with complex MOO problems, whereas our proposed one overcomes this challenge. Further discussion and analysis on the performance between Co-PSL and PSL-MOBO is presented on the Appendix.

\subsection{Ablation studies on Co-PSL}

\hl{Trade-off between two stages} Figure \ref{fig: Co_PSL_tradeoff_2stage} presents the trade-off curves between 2 stages of Co-PSL under MED and LHD metrics with a total budget of 300 expensive evaluations. On the one hand, spending more evaluations for the warm-starting stage supports Co-PSL to obtain a better and more stable Pareto front, resulting in better LHD scores. On the other hand, spending more evaluations for controllable PSL supports Co-PSL to effectively compute the optimal trade-off solution controllingly and accurately, resulting in better MED scores. By default, we spend training iterations and evaluations equally between warm-starting and controllable PSL.

\hl{Trade-off in Pareto Set Model parameter initialization} Figure \ref{fig: Co_PSL_tradeoff_beta} presents the MED and LHD curves of Co-PSL after the default 400 training iterations across different choices for trade-off scalar $\beta$ in the Pareto Set Model. Co-PSL achieved the best performances in both LHD and MED scores when $\beta$ ranged from $0.2$ to $0.4$. In our experiment, we choose $\beta = 0.2$ and gradually reduce by a factor of $10$ for every $5$ iteration, emphasizing random initialization as the training goes on.

\begin{figure}[t!]
    \centering
    \includegraphics[width = \linewidth]{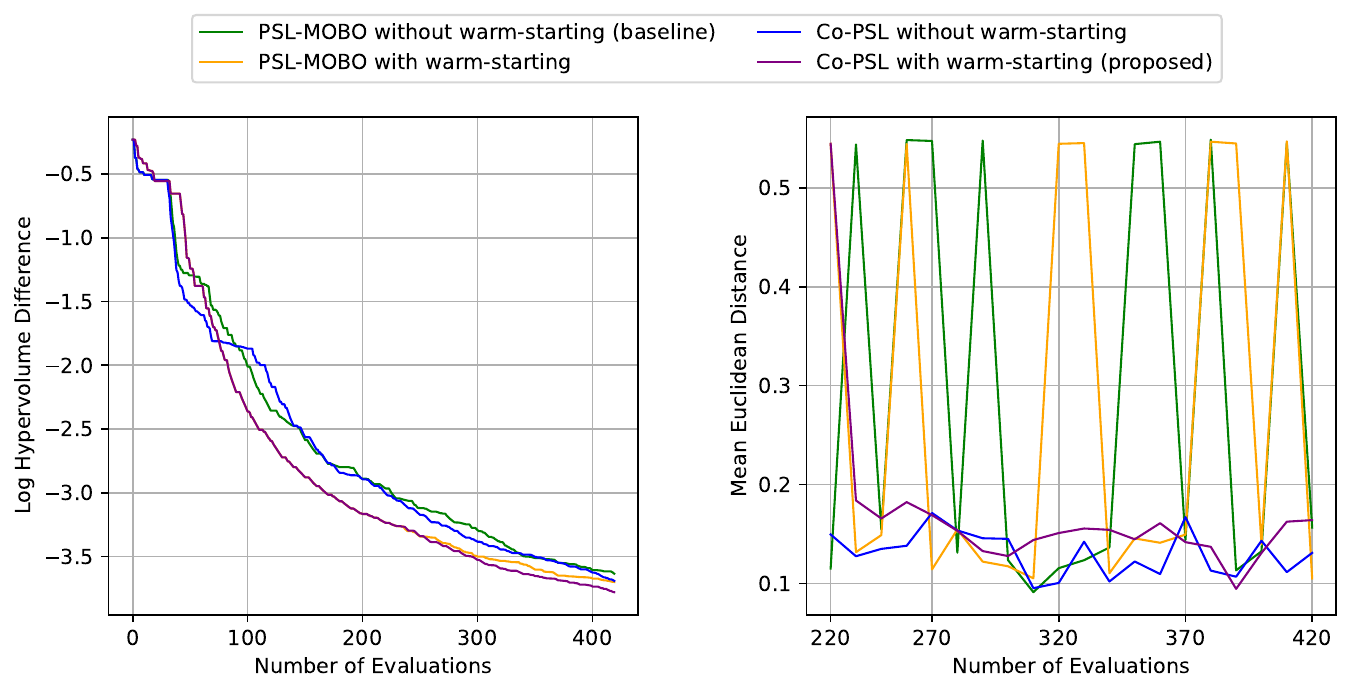}
    \caption{Performance of Co-PSL (proposed) and PSL-MOBO (baseline) on DTLZ2 with individual components on Log Hypervolume Difference and Mean Euclidean Distance.}
    \label{fig: ablation_CoPSL}
\end{figure}

\hl{Performance of individual component of Co-PSL} Figure \ref{fig: ablation_CoPSL} presents the performances of Co-PSL and PSL-MOBO with different components between the two stages, with respect to Log Hypervolume Difference (LHD) and Mean Euclidean Distance (MED). All settings are trained on 40 iterations in total with batch evaluation 10, making a total of 470 expensive evaluations eventually with 20 initial evaluations. In MED, we only plot the distance scores in the second half of the training so that equal comparisons can be made between different settings in terms of the number of expensive evaluations. Either with or without warm-starting, the main difference between Co-PSL and PSL-MOBO relies on the parameter initialization method presented in Formula (\ref{eq: theta_update}). As shown in Figure \ref{fig: ablation_CoPSL}, the warm-starting stage supports both Co-PSL and PSL-MOBO in having a better Pareto set that well approximates the truth Pareto front under LHD metric, whereas the parameter initialization method supports Co-PSL to achieve more stable and robust performance in Pareto Set learning under MED metric.

\section{Conclusion and Future Work}
\label{sec:conclusion}
% In this paper, we introduce Controllable Pareto Set Learning (Co-PSL), an effective framework for approximating the complete Pareto Front and optimizing multiple black-box objectives. Our approach consists of two main stages. Initially, we acknowledge the limitations of directly optimizing inaccurate Gaussian processes (GPs) and instead employ robust multi-objective Bayesian optimization (MOBO) techniques to gather a high-quality, expensive dataset. Subsequently, we utilize a neural network called the Pareto Set Model to learn a precise mapping between preference vectors, training it through derivative-free methods with surrogate models. Our results showcase the successful acquisition of preference mapping and a reduction in costly evaluations.

% This work also opens avenues for further research. On one hand, the evaluation of arbitrary Pareto optimal solutions generated by the Pareto Set Model remains challenging, as they are not available in the dataset obtained after the optimization processes. In real-world scenarios, the errors between the true Pareto front and the predicted one are often unknown, rendering these solutions as relative references for decision-makers. Moreover, this error becomes more pronounced in high-dimensional settings with a large number of optimized variables. On the other hand, exploring the optimal combination of algorithms for Pareto Set Learning is crucial. The efficacy of algorithms in obtaining diverse Pareto optimal solutions does not necessarily guarantee improved GP accuracy, which is essential for optimizing the Pareto Set Model.

In this paper, we introduce Controllable Pareto Set Learning (Co-PSL), an effective framework for approximating the entire Pareto Front and optimizing multiple conflicting black-box objectives. Our approach has two main stages. First, we avoid the pitfalls of directly optimizing inaccurate Gaussian processes (GPs) and instead leverage warm-starting Bayesian optimization techniques to obtain a high-quality approximation of individual objectives. Second, we train a neural network called the Pareto Set Model to learn an accurate mapping between preference vectors using derivative-free methods with surrogate models. Our results demonstrate the successful acquisition of preference mapping and a reduction in costly evaluations.

This work also suggests directions for future research. On one hand, the evaluation of arbitrary Pareto optimal solutions generated by the Pareto Set Model remains challenging, as they are not present in the dataset obtained after the optimization processes. In real-world scenarios, the errors between the true Pareto front and the predicted one are often unknown, making these solutions only relative references for decision-makers. Furthermore, this error becomes more significant in high-dimensional settings with a large number of optimized variables. On the other hand, finding the optimal combination of algorithms for Pareto Set Learning is important. The effectiveness of algorithms in obtaining diverse Pareto optimal solutions does not necessarily imply improved GP accuracy, which is vital for optimizing the Pareto Set Model.

\newpage

\newpage

%%%%%%%%%%%%%%%%%%%%%%%%%%%%%%%%%%%%%%%%%%%%%%%%%%%%%%%%%%%%%%%%%%%%%%%%%%%%%%%%%%%%%%%%%%%%%%%%%%%%%%
\appendix
\section{Instability of Pareto Set Learning in MOBO}

\begin{figure*}[!htb]
    \centering
    \begin{subfigure}[b]{0.31\textwidth}
        \centering
        \includegraphics[width=\textwidth]{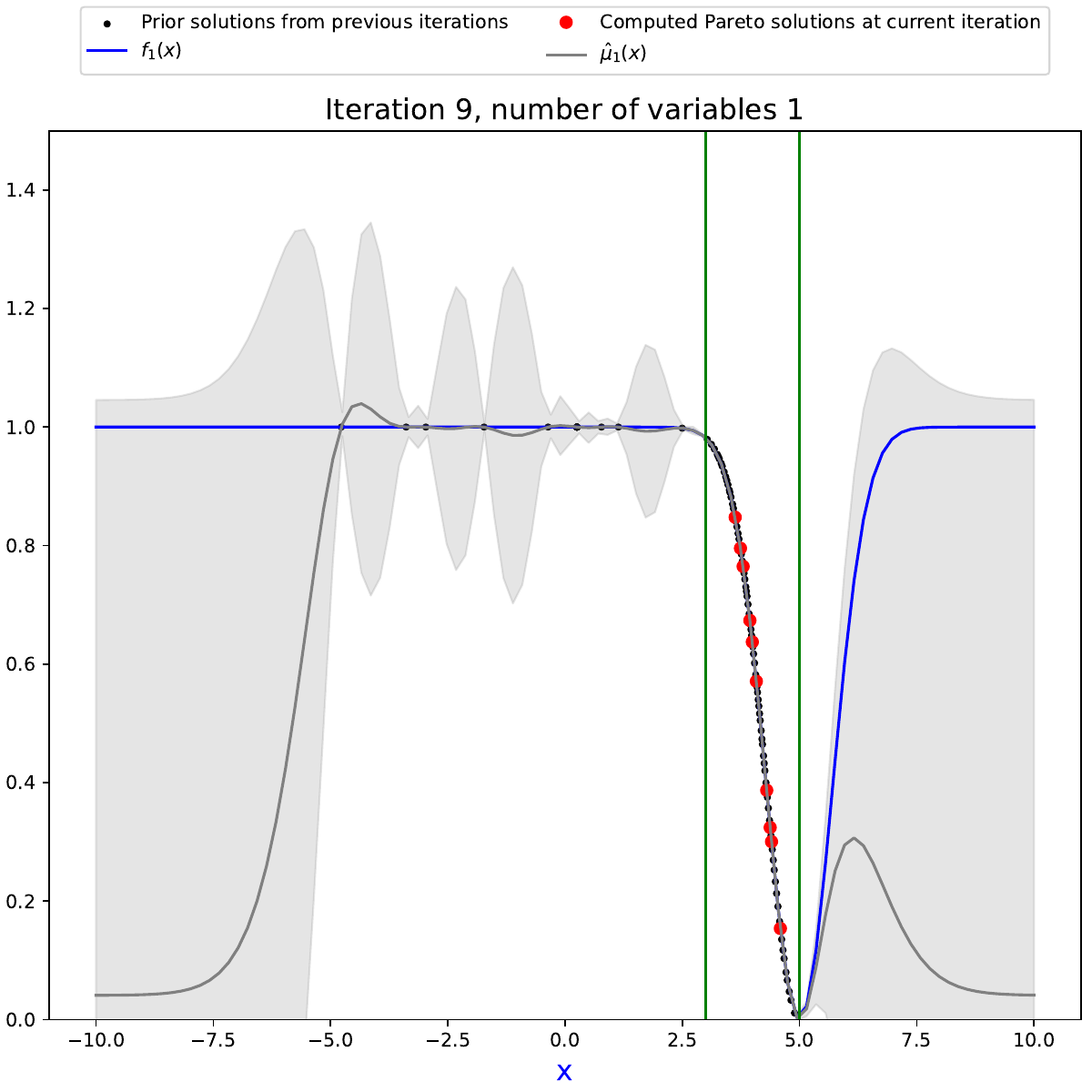}
    \end{subfigure}
    \begin{subfigure}[b]{0.31\textwidth}
        \centering
        \includegraphics[width=\textwidth]{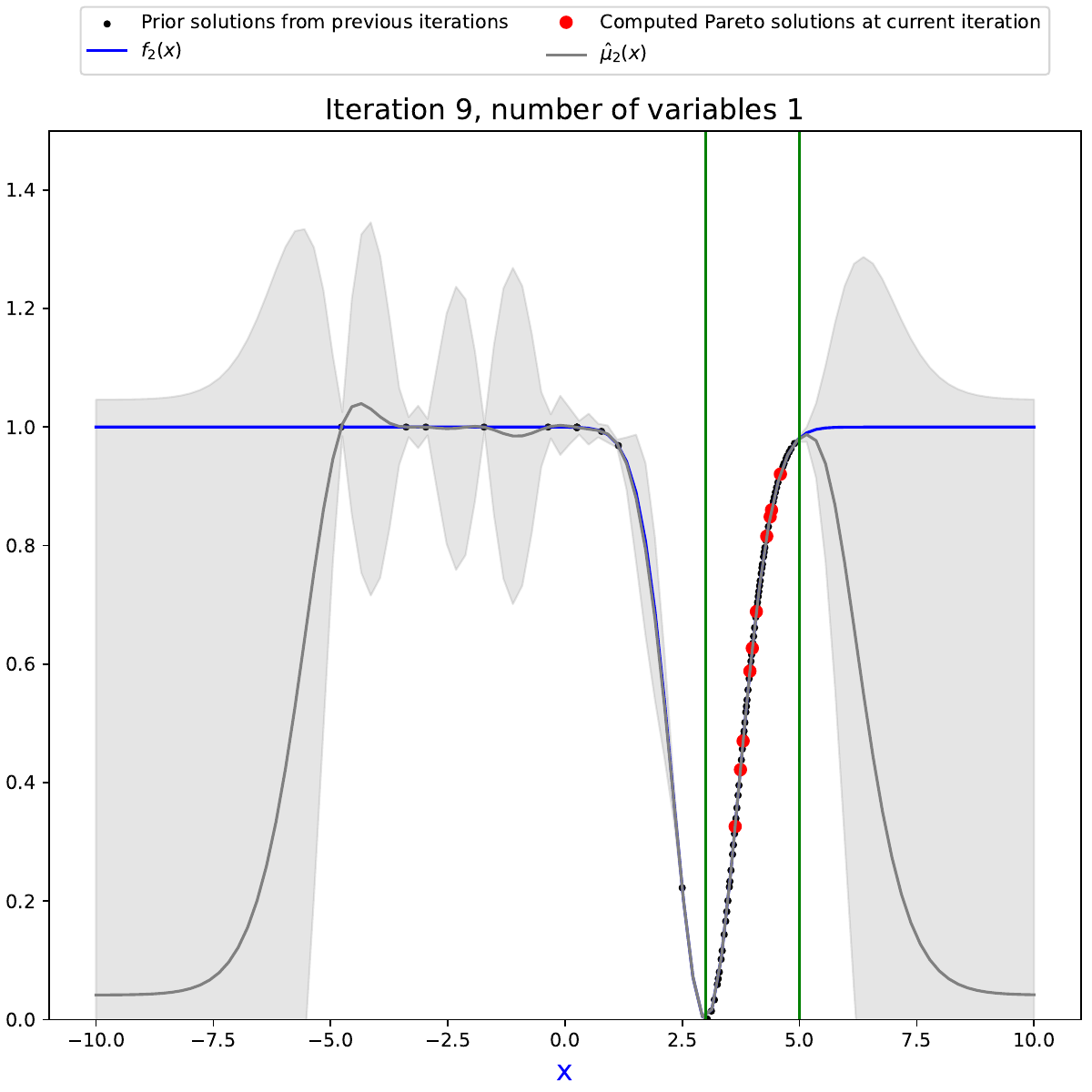}
    \end{subfigure}
    \begin{subfigure}[b]{0.31\textwidth}
        \centering
        \includegraphics[width=\textwidth]{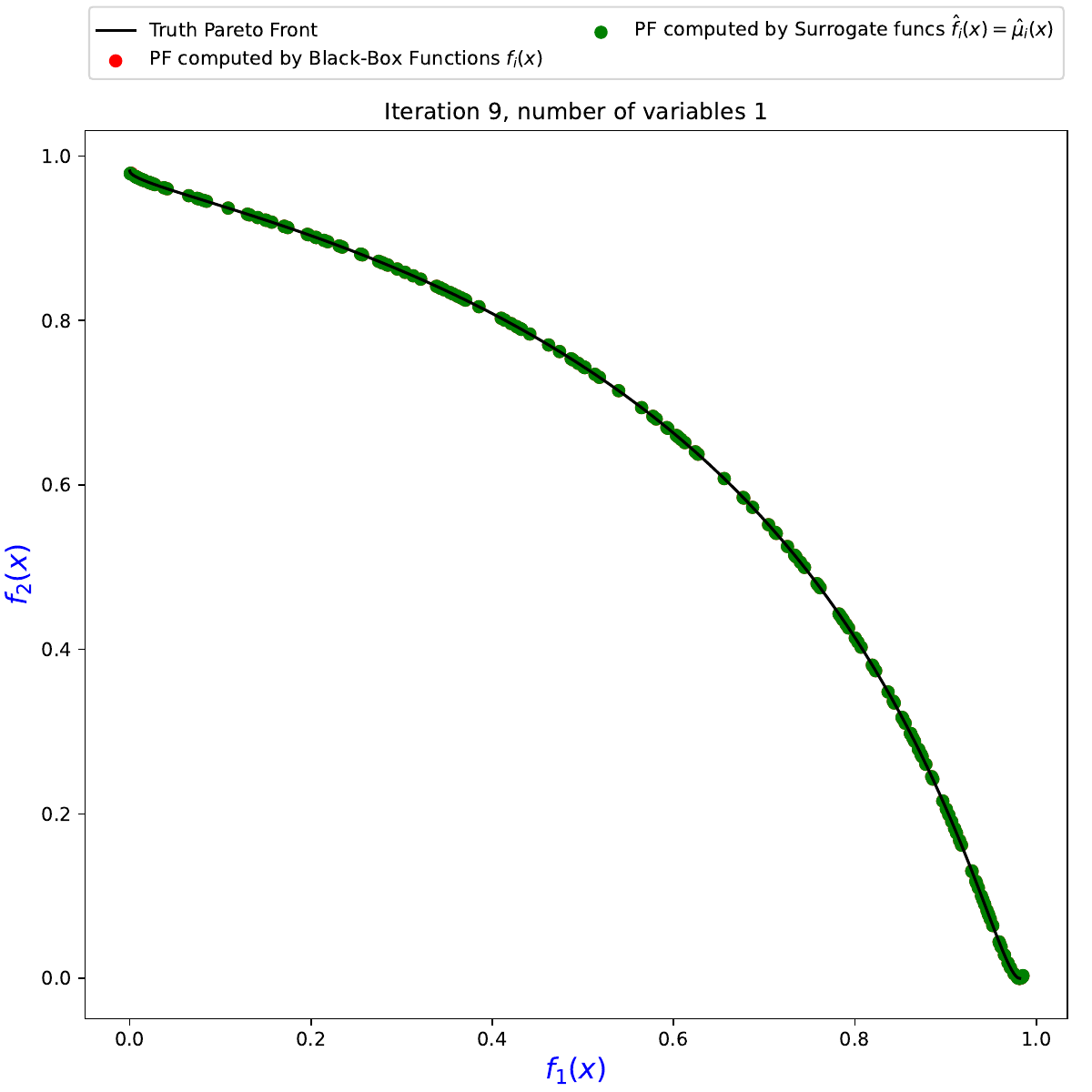}
    \end{subfigure}
    \begin{subfigure}[b]{0.31\textwidth}
        \centering
        \includegraphics[width=\textwidth]{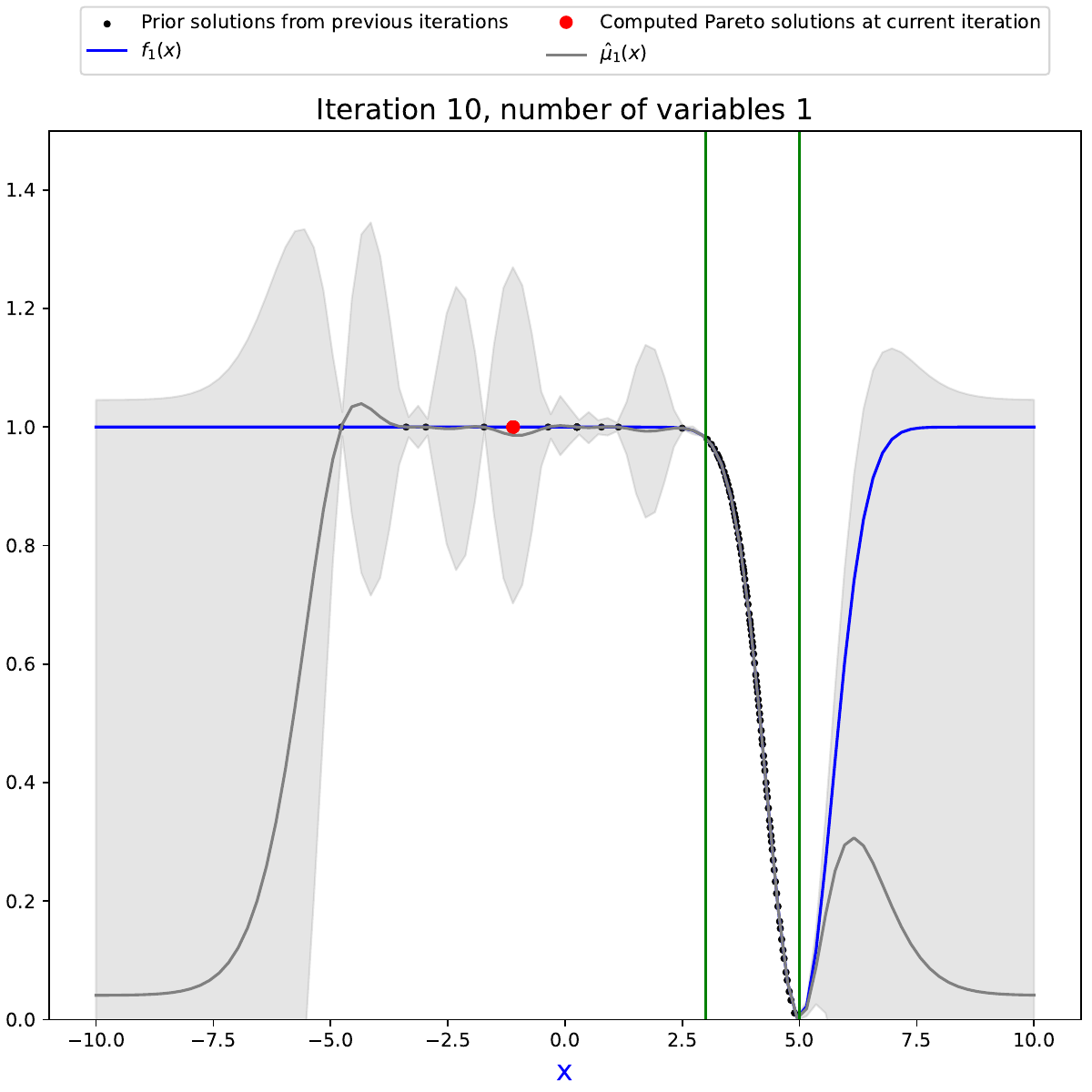}
    \end{subfigure}
    \begin{subfigure}[b]{0.31\textwidth}
        \centering
        \includegraphics[width=\textwidth]{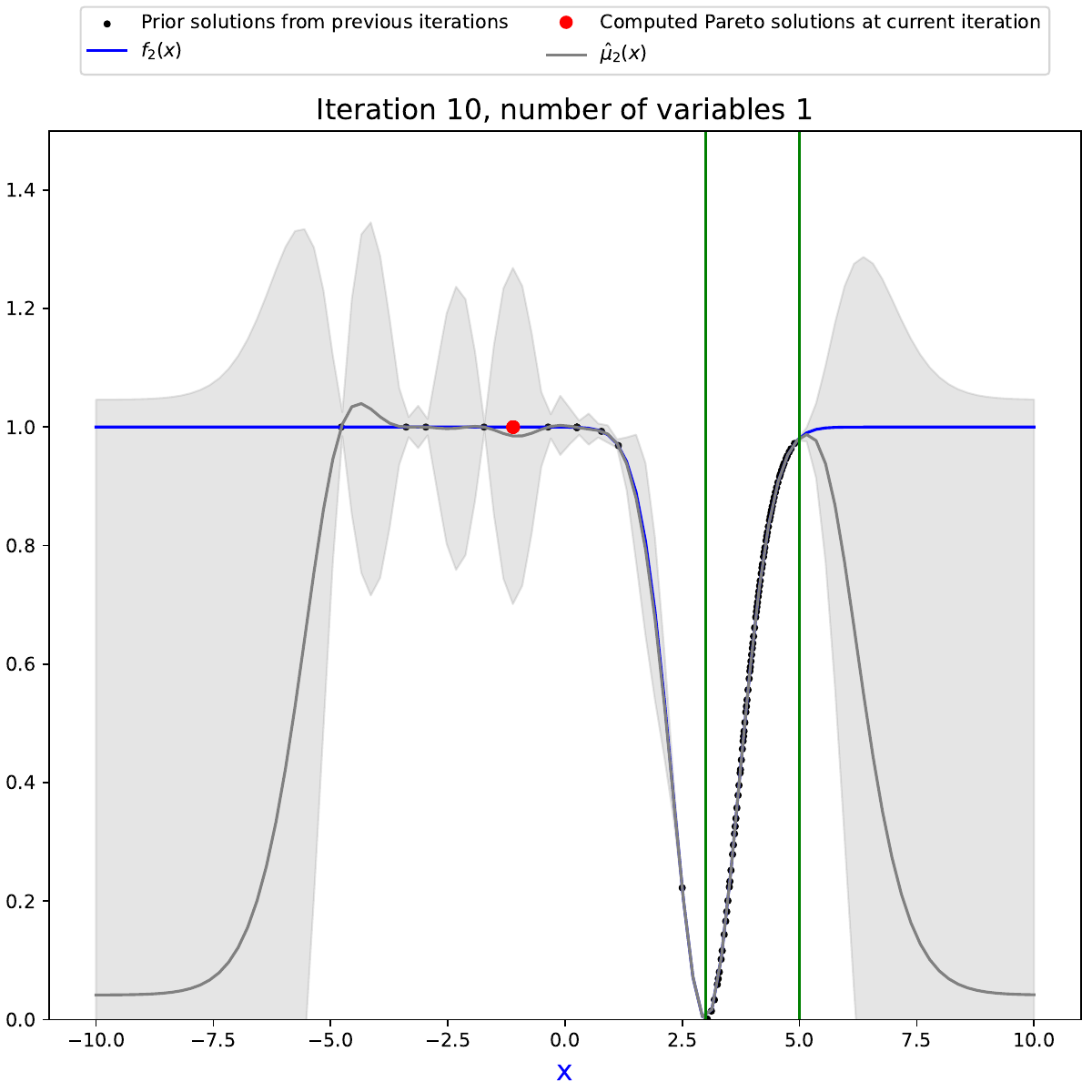}
    \end{subfigure}
    \begin{subfigure}[b]{0.31\textwidth}
        \centering
        \includegraphics[width=\textwidth]{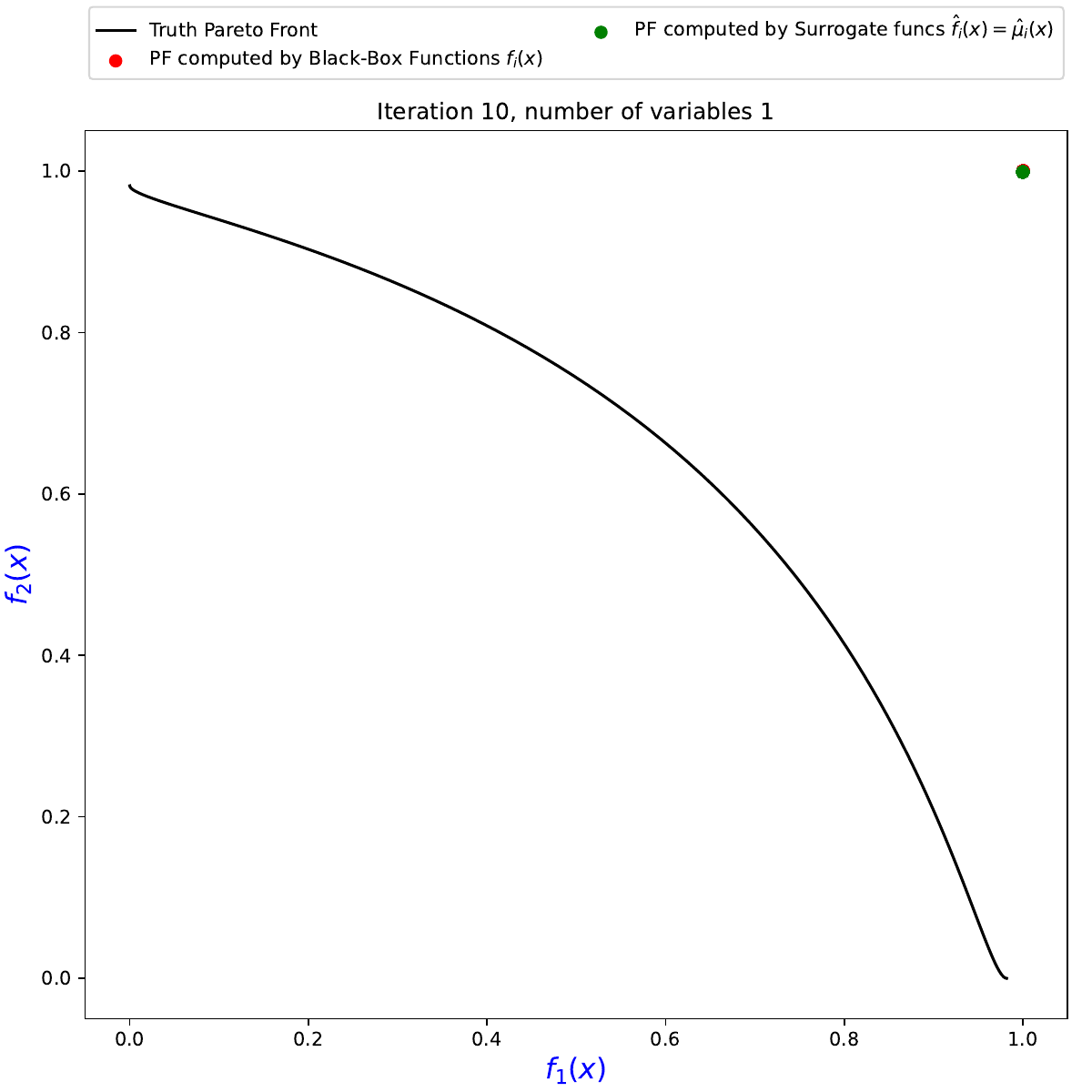}
    \end{subfigure}
     \begin{subfigure}[b]{0.31\textwidth}
        \centering
        \includegraphics[width=\textwidth]{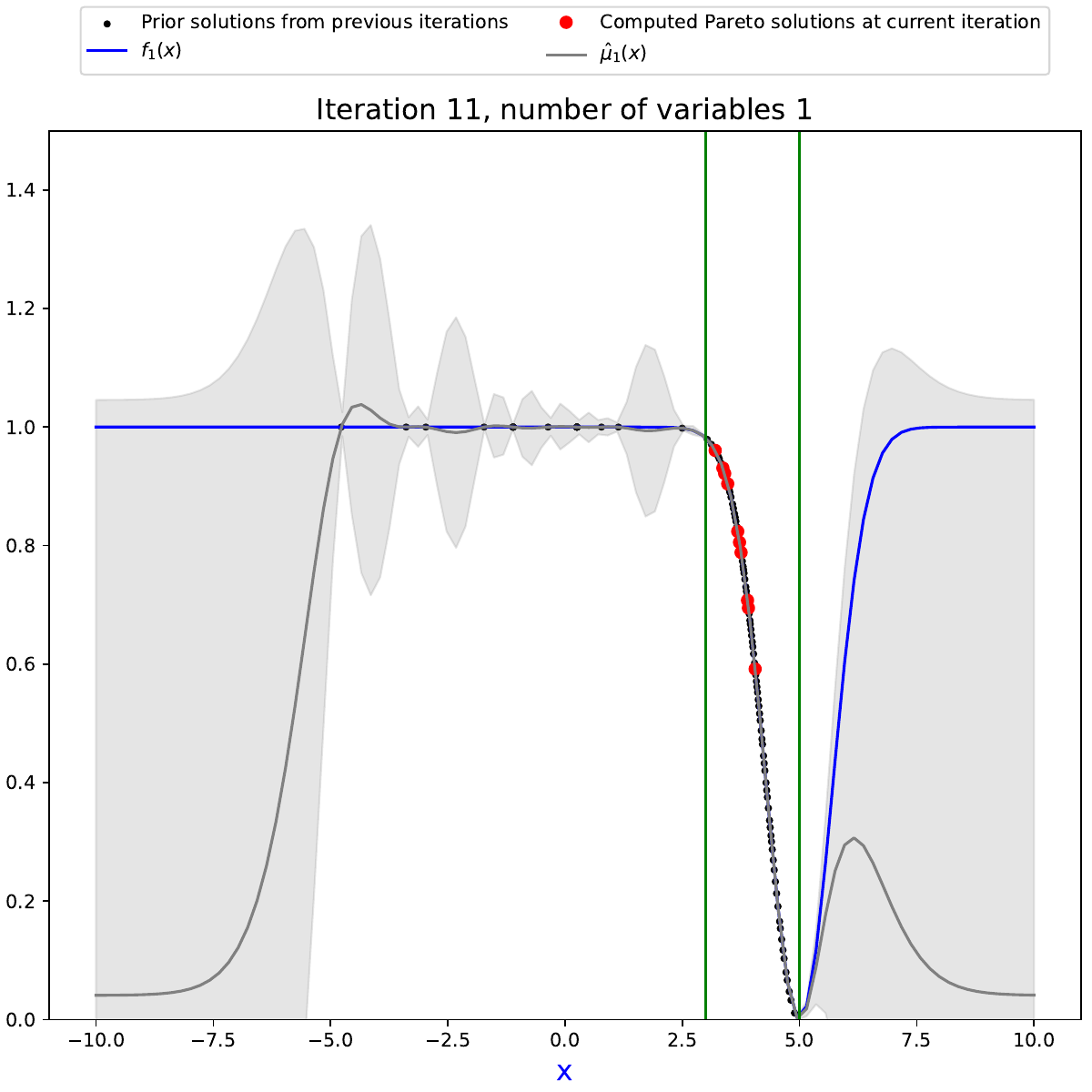}
    \end{subfigure}
    \begin{subfigure}[b]{0.31\textwidth}
        \centering
        \includegraphics[width=\textwidth]{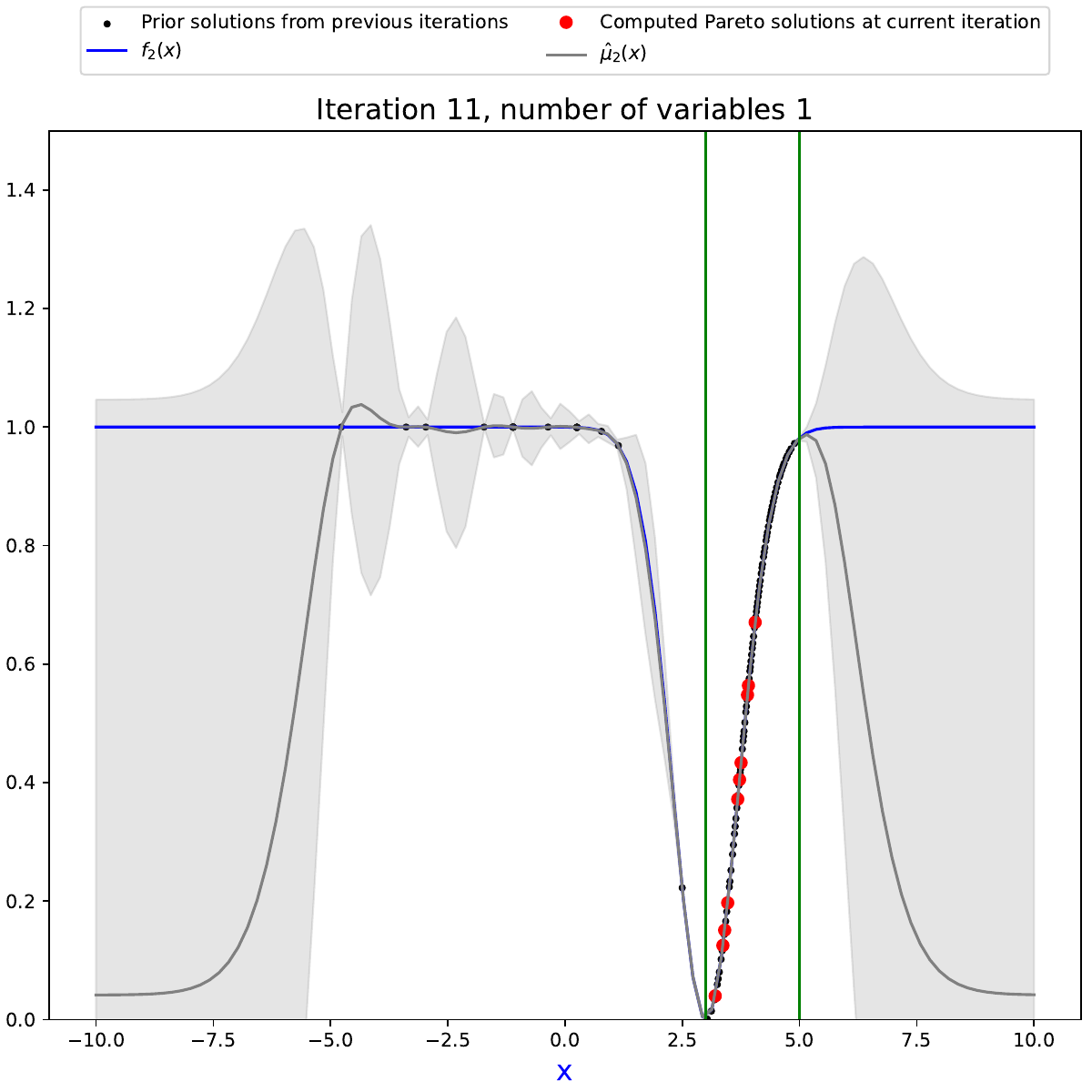}
    \end{subfigure}
    \begin{subfigure}[b]{0.31\textwidth}
        \centering
        \includegraphics[width=\textwidth]{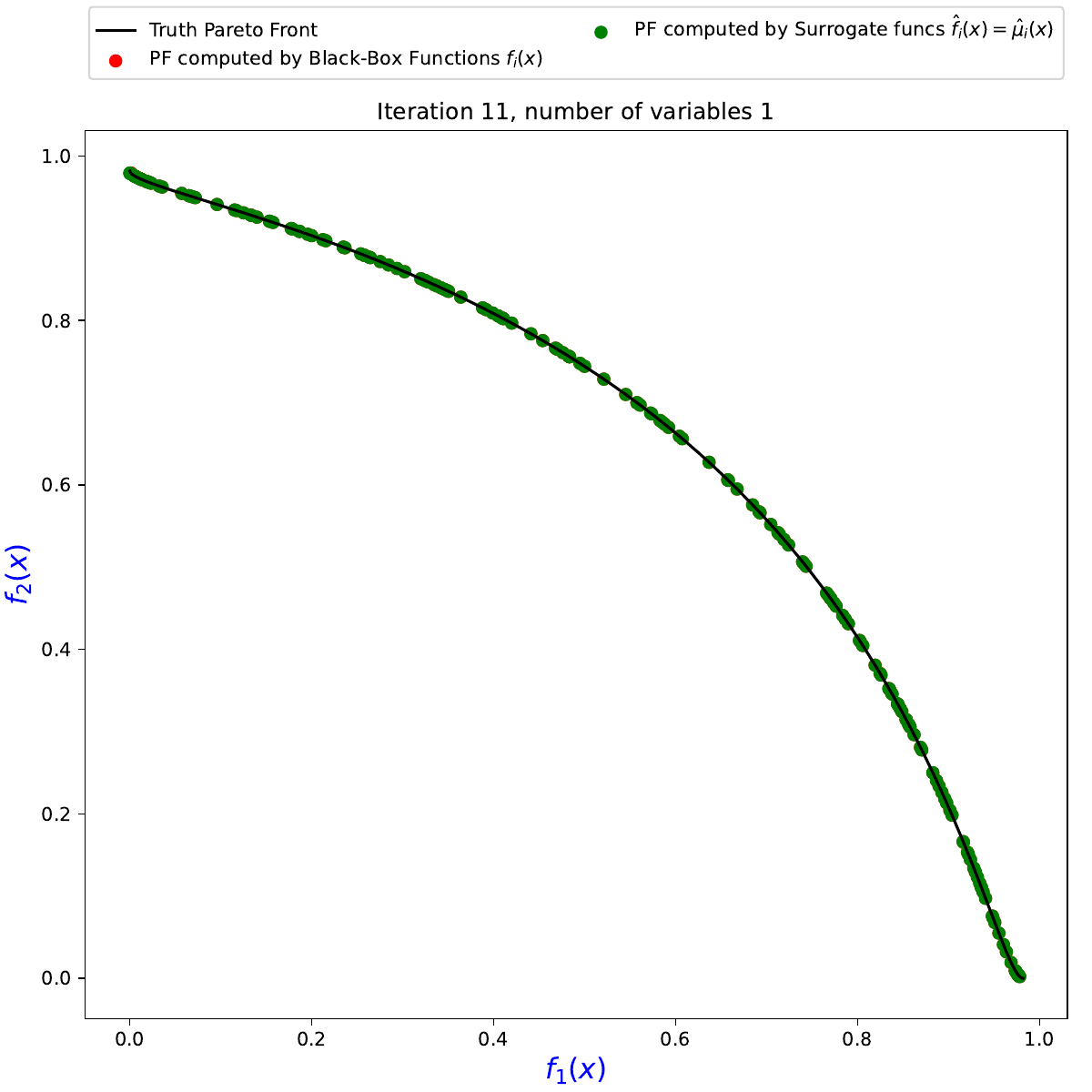}
    \end{subfigure}
    \caption{Performance of PSL-MOBO for the VLMOP problem in three consecutive iterations from 9 to 11, demonstrating the unstable performance in learning the Pareto front, which is in this case in the range $[3, 5]$ bounded by two \textcolor{green}{green vertical lines}. The \textcolor{blue}{blue line} is the truth Pareto front, while the \textit{grey line} with uncertainty region is the Gaussian Process with LCB acquisition function, approximated by the prior solutions on each iteration, represented by \textbf{black dots}. In contrast, the \textcolor{red}{red dots} are current solutions obtained after training the Pareto Set Model based on the Gaussian Process.}
    % \caption{Unstable phenomenon of Pareto Set Learning for Multi-objective Bayesian Optimization (MOBO) problem when optimizing the Pareto Set Model based on Gaussian Processes (GPs), with the Pareto Front in this case is the range $[3, 5]$, bounded by 2 green lines. At \textbf{Iteration 20} (\textit{top row}), the original algorithm PSL-MOBO performs well and the computed Pareto solutions form a good approximation of the truth Pareto Front. At the following \textbf{Iteration 21} (\textit{bottom row}), a challenge emerges due to the default initialization of the Pareto Set Model, rendering PSL-MOBO overly sensitive to inaccuracies introduced by certain portions (particularly the region around $0$ in the figures) that are poorly approximated by GPs. As a result, the algorithm falters in this particular iteration, leading to an unstable phenomenon that leads to wasted evaluations for solving the MOBO problem.}
    \label{fig: unstable_PSLMOBO}
\end{figure*}

Figure \ref{fig: unstable_PSLMOBO} demonstrates the instability performance of PSL-MOBO \citep{lin2022pareto} in a modified VLMOP2 that has the optimal region far from $0$, in this case, $[3, 5]$, across three consecutive optimization iterations 9 to 11. At interaction 9, while $h(r|\theta)$ successfully reaches the optimal region for effective mapping between $r$ and $\vf(x_r)$ that leads to a good approximation of the Pareto front, the prior solutions that were used to approximate the Gaussian Process based on LCB leave some uncertainty regions between prior solutions along with an ``pseudo-local optimal'' around the region around $x=-0.5$. For the new iteration 10, when starting at a fixed initial position, $h(r|\theta)$ has to pass through these uncertainty regions to become stuck at this particular pseudo optimal, leading to a poor $x_r$ that results in the unstable Pareto front approximation phenomenon. While this particular uncertainty region is broken down and mitigated at the following iteration 11 and removes that ``pseudo-local optimal'', another one appears around the region around $x = -2.3$, leading to a potential stuck in the next PSL round.

In contrast, our proposed Co-PSL tackles this limitation by leveraging both warm-starting Bayesian optimization and parameter re-initialization for the Pareto Set Model. While the warm-starting stage approximate black-box functions adequately to reduce the size and range of uncertainty regions, which significantly mitigate ``pseudo-local optimal'', the parameter re-initialization scheme allows $h(r|\theta)$ to begin at the stopping points from the previous iteration, avoiding facing these pseudo-optimal regions again. This results in a more stable and consistent Pareto set learning performance, as demonstrated in Figure \ref{fig: unstable_PSL}. 

\begin{table}[htpb]
    \begin{adjustbox}{max width=\linewidth}
    \centering
    \begin{tabular}{lcccr}
         \toprule
         Problem & n & m & Reference Point ($r$) & Boundary\\
         
         \midrule
         VLMOP2 & 6 & 2 & $(1.1,\ 1.1)$ & $[-2, \ 2]$\\    
         F2 & 6 & 2 & $(1.1,\ 1.1)$ & $[0, \ 1]$\\     
         DTLZ2 & 6 & 3 & $(1.1,\ 1.1,\ 1.1)$ & $[0, \ 1]$\\
         Disk Brake & 4 & 3 & $(5.8374,\ 3.4412,\ 27.5)$ & $[0, \ 1]$\\
         Gear Train  & 4 & 3 & $ (6.5241, \ 61.6,\ 0.3913)$ & $[0, \ 1]$ \\
         Rocket Injector  & 4 & 3 & $(1.0884, \ 1.0522,\ 1.0863)$ & $[0, \ 1]$ \\
         \bottomrule
    \end{tabular}
    \end{adjustbox}
    \caption{Information of synthesis and real-world problems}
    \label{tab:toy_problem_infor}
\end{table}

\section{Evaluating problem sets}
For evaluating our proposed MOBO method, we utilized widely-used synthesis multi-objective-optimization benchmark problems, including VLMOP2 \citep{van1999multiobjective} and DTLZ2 \citep{deb2002scalable} that are widely used across MOO studies, along with the second synthesis function (Function 2) from the original PSL paper \citep{lin2022pareto}. For real-world multi-objective optimization problems, we evaluated three practical problems including Disk Brake Design \citep{ray2002swarm}, Gear Train Design \citep{deb2006innovization}, Rocket Injector Design \citep{vaidyanathan2003cfd}, which can be formulated as expensive MOO problems with unknown Pareto front. Detailed descriptions of the MOO problems are described as follows:

\subsection{Synthesis problems}
\begin{itemize}
    \item \textbf{VLMOP2} with 2 objective functions:
        \begin{align*}
            &\begin{cases}
                f_1(x) = 1 - \text{exp}\left(\lVert x - \frac{1}{\sqrt{n}}\rVert_2\right) \\ 
                f_2(x) = 1 - \text{exp}\left(\lVert x + \frac{1}{\sqrt{n}}\rVert_2\right)\\
            \end{cases}\\
             \text{s.t } &x \in \bbR^6, \ -2 \leq x \leq 2. \nonumber
         \end{align*}\\
    \item \textbf{DTLZ2} with 3 objective functions:
        \begin{align*}
            &\begin{cases}
            f_1(x) = \left[1 + \lVert \vx - 0.5 \rVert_2\right]\cos(x_{1}\pi/2)cos(x_{2}\pi/2)\\
            f_2(x) = \left[1 + \lVert \vx - 0.5 \rVert_2\right]\cos(x_{1}\pi/2)sin(x_{2}\pi/2)\\
            f_3(x) = \left[1 + \lVert \vx - 0.5 \rVert_2\right]\sin(x_{1}\pi/2)\\
            \end{cases}\\
            \text{s.t } & x \in \bbR^6, \ -0 \leq x \leq 1. \nonumber
        \end{align*}
    \item \textbf{Synthesis Function 2} with 2 objective functions:
        \begin{align*}
            &\begin{cases}
                f_1(x) = (1 +\frac{s_1}{\lvert J_1 \rvert})x1\\ 
                f_2(x) = (1 +\frac{s_2}{\lvert J_2 \rvert})\left(1 - \sqrt{\frac{x_1}{1+\frac{s_2}{\lvert J_2 \rvert}}}\right)\\
            \end{cases}\\
             \text{where } & s_i= \sum_{j \in J_i}\left(x_j - \sin\left(4\pi x_1 + \frac{j\pi}{n}\right)\right)^2, \nonumber\\
             \text{and }& J_1 = \{j \mid j \mathrel{\not \vdots} 2 ;\ j \in [2, \ n]\}, \nonumber\\ 
                        &J_2 = \{j \mid j \mathrel{\vdots} 2; \ j \in [2, \ n]\}, \nonumber\\
             \text{s.t } & x \in \bbR^6, \ -0 \leq x \leq 1. \nonumber
        \end{align*}
\end{itemize}

\subsection{Real-world application problems}
\begin{itemize}
    \item \textbf{Disk Brake Design} This problem aims to design disk brakes, with four decision variables including the inner radius, the outer radius, the engaging force, and the number of friction surfaces. The target of this problem is minimizing three objectives including the mass, the stopping time, and the violations of four designs contained, with the detailed description presented in \citep{ray2002swarm}
    
    \item \textbf{Gear Train Design} This problem aims to design a gear train with four gears, where the decision variables are the number of teeth in each gear. The target of this problem is maximizing the size of four gears, minimizing the design-constrained violations, and minimizing the differences between the realized gear ratio and the required specification. Detailed of this problem is presented in \citep{deb2002scalable}
    
    \item \textbf{Rocket Injector Design} This problem aims to design a rocket injector with four variables: the hydrogen flow angle, the hydrogen area, the oxygen area, and the oxidizer post-tip thickness. The objective is to minimize the maximum temperature of the injector face, the distance from the inlet, and the temperature on the post tip. Detailed of this problem is in \citep{vaidyanathan2003cfd}

\end{itemize}

The three real-world multi-objective problems do not have exact Pareto fronts, leaving these problems truly black-box. For evaluation, we use the approximate Pareto fronts provided by \citep{tanabe2020easy} as ground truth for evaluating these problems. We used the reference point for computing the hypervolume by $r = 1.1 \times z_{nadir}$, where $z_{nadier}$ is the nadir point of the truth Pareto front. A summary of these problems is presented in Table \ref{tab:toy_problem_infor}.

% \begin{figure*}[!ht]
%     \centering
%     \begin{subfigure}[b]{0.4\textwidth}
%         \centering
%         \includegraphics[width=\textwidth]{Images/Front_DTLZ2-PSL-MOBO.pdf}
%     \end{subfigure}
%     \begin{subfigure}[b]{0.4\textwidth}
%         \centering
%         \includegraphics[width=\textwidth]{Images/Front_DTLZ2-Co-PSL.pdf}
%     \end{subfigure}
%     \caption{Difference between the truth and the learned Pareto Fronts and its corresponding Mean Euclidean Distance scores between PSL-MOBO and Co-PSL on DTLZ2 with 6 variables.}
%     \label{fig: DTLZ2_front}
% \end{figure*}

% \begin{figure}[htbp]%
%     \centering
%     \subfloat[\centering Performances of Co-PSL across warm-up methods]{{\includegraphics[width=\textwidth]{Images/Co-PSL_ablation_warmup.pdf}}\label{fig: Co_PSL_warmup}}
%     \quad
%     \subfloat[\centering Performances of Co-PSL across initialization methods]{{\includegraphics[width=\textwidth]{Images/Co-PSL_ablation_noise.pdf}}\label{fig: Co_PSL_noise}}
%     \caption{Ablation studies for Co-PSL on DTLZ2 with Log Hypervolume Difference (left column) and Mean Euclidean Distance (right column) between (a) different choices for warm-starting methods and numbers of batch evaluation and (b) difference choices for parameter initialization for Pareto set model.}
% \end{figure}

\section{Co-PSL configurations}

% \subsection{Illustrations of the learned Pareto Front across PSL training}

% Figure \ref{fig: DTLZ2_front} illustrates the differences between the computed Pareto Front $\hat{\mathcal{P}} = \{ \hat{\mu}(h(r^k|\theta)), r^k \in \mathbf{r}\}$ and their corresponding truth Pareto Front $\mathcal{P} = \{ \vf(\vx^*_{r^k}), r^k \in \mathbf{r}\}$ under the same set of trade-off preference vectors $\mathbf{r} = \{r^k\}_{k=1}^K$ and their respective Mean Euclidean Distance score. Here, $K = 300$ preference vectors are sampled evenly distributed in the preference spaces using the method proposed by \cite{das1998normal}. The front is computed after training Pareto Set Model $h(r|\theta)$ with the Gaussian Process approximated by from 260 to 420 expensive evaluations (equivalent to training Pareto Set Model with training iterations $2 \leq b \leq 20$ with batch evaluations $b=10$ after having prior GPs with 220 evaluations). The Pareto Fronts learned by PSL-MOBO are highly unstable and inaccurate, as explained in detail in Figure 1 of the main manuscript, whereas Co-PSL provides a more consistent and accurate Pareto Front across training iterations and evaluations.

\subsection{Choosing GP optimization method for warm-starting stage}

\begin{figure}[htbp]
    \centering
    \includegraphics[width = .9\linewidth]{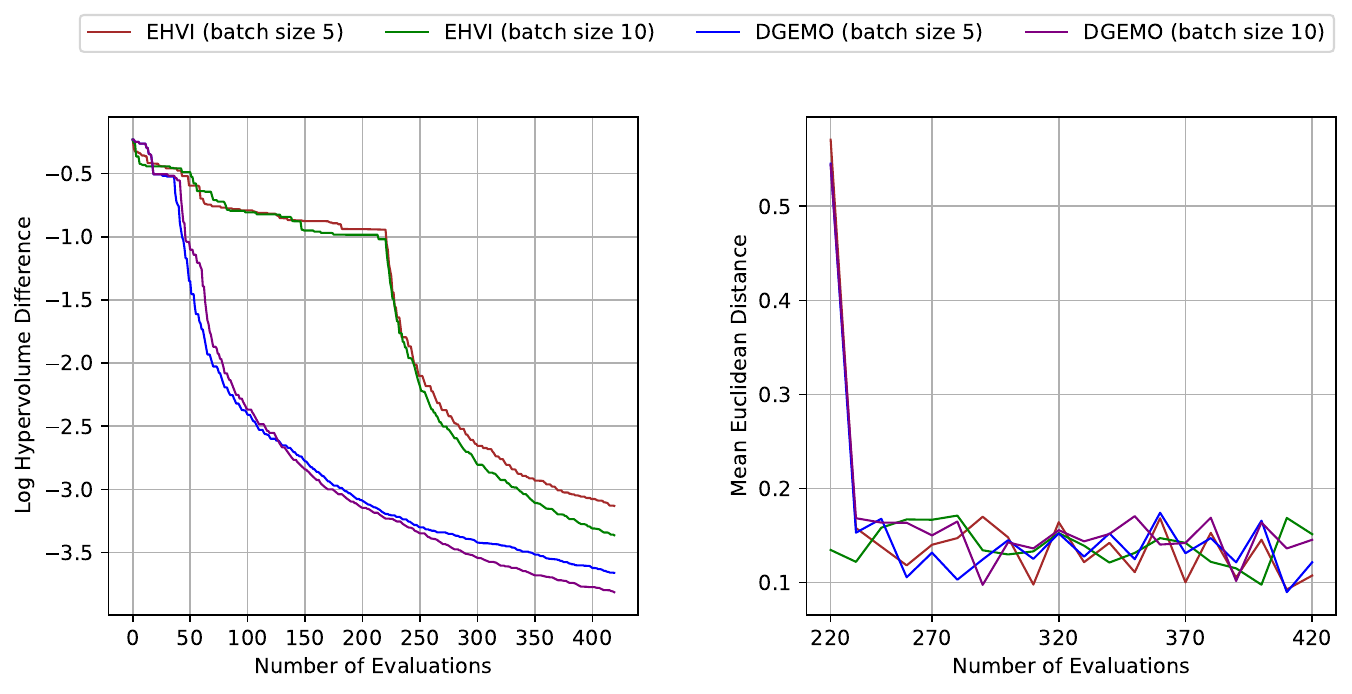}
    \caption{Performances of Co-PSL across warm-up methods.}
    \label{fig: Co_PSL_warmup}
\end{figure}

For constructing the warm-starting Bayesian optimization (BO), we utilized the Gaussian process optimized based on Expected Hypervolume Improvement (EHVI) and Diversity-Guided Efficient Multi-Objective Optimization (DGEMO) as the BO-based warm-starting methods. Furthermore, we trained 2 stages of Co-PSL with batch evaluation either with $b = 5$ or $b = 10$, corresponding with $n= 40$ and $n = 20$. The LHD and MED score plots are presented in Figure \ref{fig: Co_PSL_warmup}. We recognized that DGEMO performed better under LHD and was more stable under MED compared with EHVI. Furthermore, having a larger number of batch evaluations and reducing the number of training iterations, respectively, not only improves the overall performance of Co-PSL but also saves optimizing time significantly, as the cost of training GPs per iteration step is extremely costly. Therefore, we eventually chose DGEMO for the first stage of Co-PSL, with a batch size of $b = 10$ and a number of iterations of $n = 20$ for both stages of Co-PSL.

\subsection{Choosing the initialization method for Pareto Set Model}

\begin{figure}[htbp]
    \centering
    \includegraphics[width = .9\linewidth]{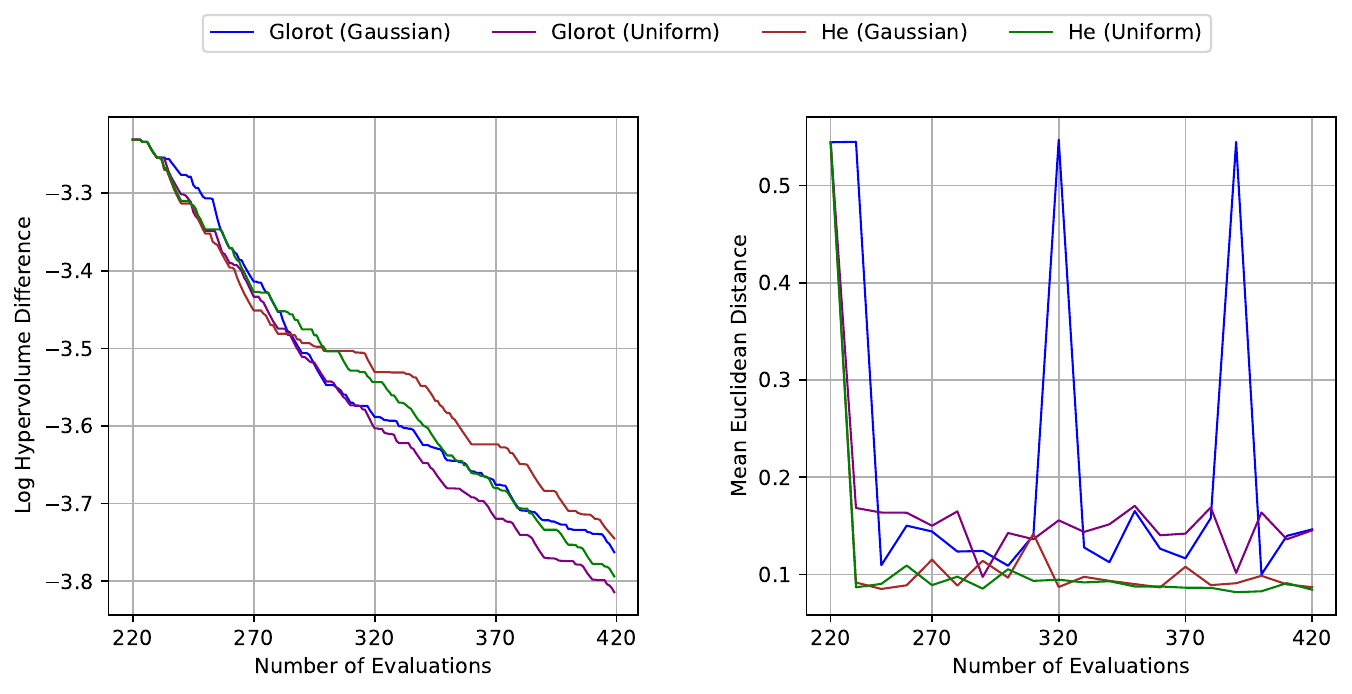}
    \caption{Performances of Co-PSL across initialization methods.}
    \label{fig: Co_PSL_noise}
\end{figure}

In the Controllable PSL stage, we implement Glorot initialization \cite{glorot2010understanding} (also known as Xavier initialization) and He initialization \cite{he2015delving} either in Normal distribution or Unifrom distribution for parameter initialization for Pareto Set Model as in Formula 8 of the main manuscript. The LHD and MED score plots across Co-PSL with these initialization methods are presented in Figure \ref{fig: Co_PSL_noise}. We only plot the LHD scores for the second stage of Co-PSL, since the first stage shares the same expensive warm-starting solutions. On the one hand, Uniform distribution supports Co-PSL to achieve better LHD scores and more stable MED scores compared to Normal distribution. On the other hand, He initialization supports Co-PSL to perform better in terms of MED score. Eventually, we decided to use Glorot initialization with Uniform distribution as the method for parameter initialization.

\bibliography{reference}

\end{document}